\documentclass{article}

\PassOptionsToPackage{numbers, compress}{natbib}
\usepackage[preprint]{dlwlrma_2026}

\usepackage[utf8]{inputenc} % allow utf-8 input
\usepackage[T1]{fontenc}    % use 8-bit T1 fonts
\usepackage{hyperref}       % hyperlinks
\usepackage{url}            % simple URL typesetting
\usepackage{booktabs}       % professional-quality tables
\usepackage{amsfonts}       % blackboard math symbols
\usepackage{nicefrac}       % compact symbols for 1/2, etc.
\usepackage{microtype}      % microtypography
\usepackage{xcolor}         % colors
\usepackage{multirow}
\usepackage{graphicx}
\usepackage{array}
\usepackage{wrapfig}
\usepackage{placeins}
\usepackage{amsmath}
\usepackage{amssymb}
\title{Continuous Expert Assembly: Instance-Conditioned Low-Rank Residuals for All-in-One Image Restoration}

\author{%
  Haisen He \\
  Southern University of Science and Technology \\
  \And
  Xiangyu Zou \\
  Shenzhen University \\
  \AND
  SongLin Dong \\
  Shenzhen University of Advanced Technology \\
  \And
  Heng Li \\
  Shenzhen University of Advanced Technology \\
  \And
  Yihong Gong \\
  Xi'an Jiaotong University \\
  \And
  Zhiheng Ma\thanks{Corresponding author} \\
  Shenzhen University of Advanced Technology \\
}

\begin{document}

\maketitle

\begin{abstract}
Real-world image degradation is often unknown, spatially non-uniform, and compositional, requiring all-in-one restoration models to adapt a single set of weights to diverse local corruption patterns without test-time degradation labels. Existing methods typically modulate a shared backbone with global prompts or degradation descriptors, or route features through predefined expert pools. However, compact global conditioning can bottleneck localized degradation evidence, while static expert routing may produce homogeneous updates or rely on unstable sparse assignments. We propose \textbf{Continuous Expert Assembly} (CEA), a token-wise dynamic parameterization framework for all-in-one image restoration. CEA employs a lightweight \textbf{Cross-Attention Hyper-Adapter} to probe intermediate spatial features and synthesize instance-conditioned low-rank routing bases and residual directions. Each spatial token then assembles its own residual update via dense signed dot-product affinities over the generated rank-wise components, avoiding external prompts, static expert banks, and discrete Top-$K$ selection. The resulting assembly rule also admits a linear-attention perspective, making its dense token-wise routing behavior transparent. Experiments on AIO-3, AIO-5, and CDD-11 show that CEA improves average restoration quality over strong prompt-, descriptor-, and expert-based baselines, with the clearest gains on spatially varying and compositional degradations, while maintaining favorable parameter, FLOP, and runtime efficiency.
\end{abstract}

\section{Introduction}
\label{sec:intro}

Real-world photographs rarely suffer from one form of degradation in isolation.
Low light, haze, rain, motion blur, and sensor noise often co-occur in a single
image, are spatially non-uniform, and are unknown at inference time. All-in-one
(AiO) image restoration~\citep{li2022airnet,potlapalli2023promptir,guo2024onerestore}
therefore requires a single model to handle multiple degradation types and their
compositions without test-time knowledge of the corruption. While
Transformer-based restorers~\citep{liang2021swinir,wang2022uformer,zamir2022restormer,chen2022cat}
have substantially advanced single-degradation performance through long-range
and multi-scale modeling, AiO introduces a distinct difficulty: one set of weights
must serve heterogeneous, spatially non-uniform corruptions that often appear in composition.

Existing AiO restorers commonly adapt a shared backbone in one of two ways.
The first uses global conditioning signals, such as degradation
representations, task priors, learned prompts, language instructions, or
explicit degradation descriptors
~\citep{li2022airnet,liu2022tape,zhang2023idr,potlapalli2023promptir,conde2024instructir,guo2024onerestore}.
These signals are compact and effective, but they can compress spatially
heterogeneous degradation evidence into an image-level representation. The
second allocates conditional capacity through expert branches or low-rank
parameter bases
~\citep{shazeer2017moe,fedus2022switch,lepikhin2020gshard,puigcerver2024softmoe,guo2024adaptir,zamfir2025moceir,serrano2024abair,ai2024lorair,cao2024hair}.
Such designs provide specialization, but they often rely on a fixed expert pool
or sparse Top-$K$ routing, which can introduce assignment instability,
capacity-handling overhead, or homogeneous updates across spatial locations.
What remains missing is a mechanism that is both input-adaptive and spatially
adaptive, without collapsing the image into a global descriptor or restricting
adaptation to static expert parameters.

We argue that AiO restoration requires a more flexible form of parameter
adaptation. Instead of selecting from a static set of experts or modulating
the network with a single global condition, the model should dynamically
assemble restoration parameters conditioned on the input image and apply them
at the token level. This formulation is particularly suitable for
compositional degradations, where different spatial tokens can receive different
residual updates while sharing a compact and efficient input-conditioned
parameter space.

\begin{figure}[h]
    \centering
    \includegraphics[width=1.0\textwidth]{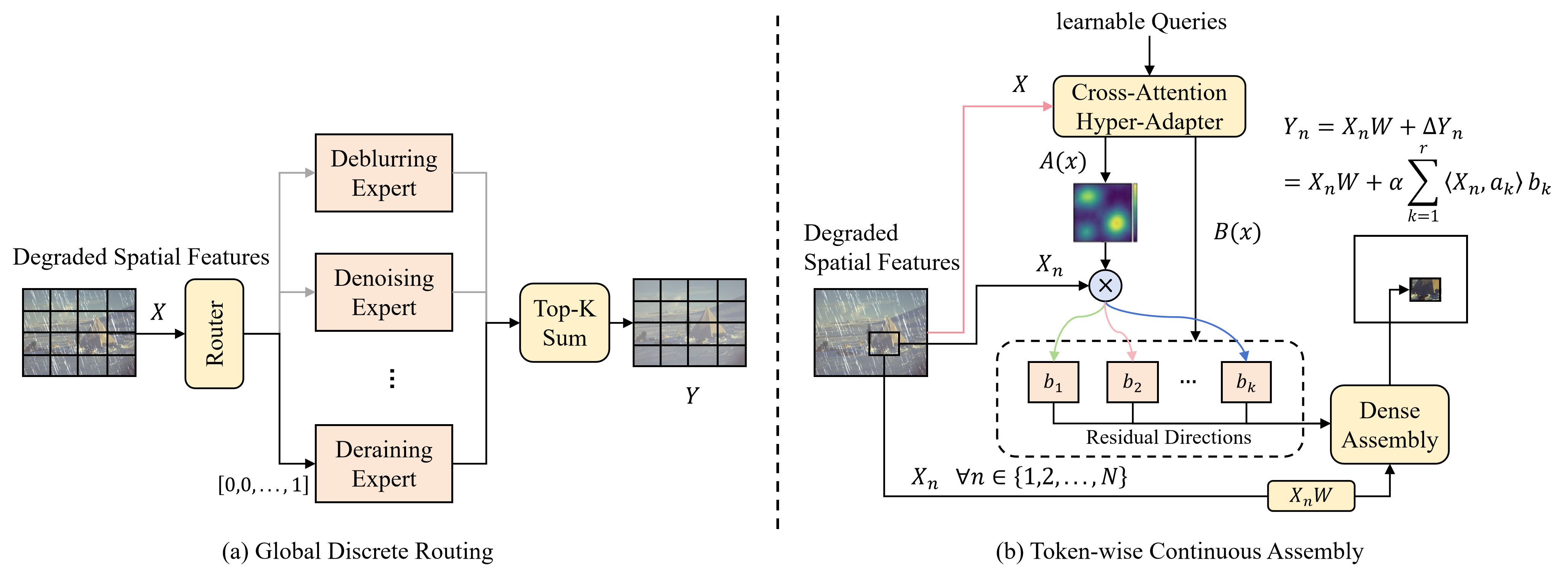}
    \caption{\textbf{Conceptual comparison of adaptation paradigms for All-in-One Image Restoration.} 
    (a) Representative global or discrete expert adaptation. A compact image-level signal or sparse router selects or modulates a fixed set of parameters, which can become coarse when degradations vary spatially or appear in composition.
    (b) Continuous Expert Assembly (CEA). A lightweight hyper-adapter probes spatial features and generates instance-conditioned low-rank routing bases \(A(x)\) and residual directions \(B(x)\). Each spatial token uses dense signed dot-product affinities to assemble its own residual update \(\Delta Y_n\), enabling instance-conditioned low-rank residual assembly without discrete expert selection.
    }
    \label{fig:1} 
\end{figure}

To this end, we propose \textbf{Continuous Expert Assembly} (CEA), a token-wise
dynamic parameterization framework for all-in-one image restoration. CEA
synthesizes instance-conditioned low-rank routing bases and residual directions
from the current degraded image. Each spatial token then assembles its own
additive residual update through dense signed dot-product affinities over these
generated rank-wise components. In this way, CEA replaces discrete selection
from fixed experts with continuous assembly in an input-generated residual
subspace, enabling fine-grained adaptation without static expert pools or hard
routing decisions.

CEA is implemented with a lightweight \textbf{Cross-Attention Hyper-Adapter}.
Rather than pooling intermediate features into a single global descriptor, the
adapter uses learnable queries to probe condensed spatial contexts and decode
them into routing bases and residual directions. The assembled residual updates
are injected into the attention projections of the restoration backbone,
allowing spatial feature interactions to adapt to the degradation structure of
each image.

From a modeling perspective, the proposed assembly rule admits a linear-attention-style algebraic view. The token-wise residual can be written as a
dot-product interaction between spatial tokens and dynamically synthesized
routing bases, followed by a weighted combination of synthesized residual
directions. In this form, spatial tokens serve as queries, while the generated
routing bases and residual directions play roles analogous to keys and values.
This view makes the dense routing behavior of CEA transparent: each token
continuously assembles its own residual update from an input-conditioned
low-rank component space.

We evaluate CEA on standard AiO restoration benchmarks, including AIO-3,
AIO-5, and CDD-11 compositional degradations. CEA improves average restoration quality over prompt-, descriptor-, and expert-based baselines, with the clearest gains under spatially varying and compositional degradations, while maintaining favorable parameter, FLOP, and runtime efficiency.

Our contributions are summarized as follows:
\begin{itemize}
    \item We propose \textbf{Continuous Expert Assembly}, a token-wise dynamic
    parameterization framework that synthesizes input-conditioned low-rank
    routing bases and residual directions for fine-grained AiO restoration.

    \item We design a lightweight \textbf{Cross-Attention Hyper-Adapter} that
    probes spatial degradation contexts and generates dynamic low-rank factors
    on the fly, avoiding both global prompt bottlenecks and static expert
    pools.

    \item We present a linear-attention-style interpretation of CEA, showing that
dense token-wise residual assembly can be written as softmax-free dot-product
aggregation between spatial tokens, generated routing bases, and generated
residual directions.

    \item Extensive experiments on AIO-3, AIO-5, and CDD-11 demonstrate that
    CEA achieves strong average restoration performance with favorable
    parameter count, computational cost, and runtime efficiency, especially
    under compositional degradations.
\end{itemize}

\section{Related Work}

\noindent\textbf{Restoration backbones.}
Image restoration has evolved from discriminative CNN restorers such as
DnCNN~\citep{zhang2017dncnn}, MPRNet~\citep{zamir2021mprnet},
HINet~\citep{chen2021hinet}, and NAFNet~\citep{chen2022nafnet}, to
architectures with stronger global interaction, including IPT~\citep{chen2021ipt}
and MAXIM~\citep{tu2022maxim}. Transformer-based restorers have further improved
long-range dependency modeling and high-resolution restoration through
efficient attention and hierarchical designs, as in SwinIR~\citep{liang2021swinir},
Uformer~\citep{wang2022uformer}, Restormer~\citep{zamir2022restormer}, and
CAT~\citep{chen2022cat}. These backbones provide strong restoration priors,
but they are typically optimized for individual degradations. AiO restoration
requires a single backbone to adapt its computation under unknown, spatially varying, and compositional degradations.

% \paragraph{All-in-One image restoration.}
\noindent\textbf{All-in-one restoration and conditioning.}
AiO restoration targets a single model that handles multiple degradation types under unknown corruption. 
AirNet learns degradation representations to guide a shared restorer~\citep{li2022airnet}; TAPE introduces task-agnostic prior embeddings
for cross-task restoration~\citep{liu2022tape}; IDR formulates multi-degradation
restoration through reusable degradation ingredients~\citep{zhang2023idr}; and
TransWeather addresses unified adverse-weather restoration~\citep{valanarasu2022transweather}.
A related line conditions restoration with learned prompts, visual prompts,
language instructions, or explicit degradation descriptors, including
PromptIR~\citep{potlapalli2023promptir}, ProRes~\citep{ma2023prores},
PIP~\citep{li2023pip}, InstructIR~\citep{conde2024instructir}, and
OneRestore~\citep{guo2024onerestore}. These methods are effective when the
conditioning signal captures the degradation well, but global prompts or
descriptors can become coarse when degradations vary spatially or appear in
composition. CEA instead generates parameter components from spatially probed
features and assembles token-specific updates.

% \paragraph{Conditional computation and expert allocation.}
\noindent\textbf{Conditional computation and expert allocation.}
Mixture-of-Experts allocates capacity by routing inputs to expert modules~\citep{shazeer2017moe}. Scalable sparse MoE systems such as Switch Transformer and GShard activate only a subset of experts and typically introduce routing decisions, capacity handling, or load-balancing objectives
~\citep{fedus2022switch,lepikhin2020gshard}. 
Vision MoE extends this idea to visual recognition~\citep{riquelme2021vmoe}. 
In image restoration, AdaptIR uses heterogeneous expert branches for parameter-efficient adaptation~\citep{guo2024adaptir}, while MoCE-IR introduces complexity experts for all-in-one restoration~\citep{zamfir2025moceir}. 
Soft MoE replaces hard Top-\(K\) assignment with dense differentiable mixing over a fixed expert set~\citep{puigcerver2024softmoe}. 
CEA shares the goal of avoiding hard expert assignment, but differs in the object being mixed. Sparse and soft MoE methods mix outputs or representations produced by a sample-independent expert bank. In contrast, CEA first generates an input-conditioned low-rank residual subspace and then lets each spatial token assemble an additive update over this generated subspace. Thus, the rank-wise components are not fixed expert networks, and the routing weights are signed dot-product affinities rather than normalized probabilities over a static expert pool.

\noindent\textbf{Low-rank and dynamic parameterization.}
Parameter-efficient adaptation methods, including Adapters~\citep{houlsby2019adapters}, LoRA~\citep{hu2022lora}, AdaLoRA~\citep{zhang2023adalora}, and DoRA~\citep{liu2024dora}, show that low-rank updates can efficiently adapt large backbones. 
Hypernetworks and dynamic filters generate sample-conditioned parameters for adaptive computation~\citep{ha2016hypernetworks,jia2016dynamic,
yang2019condconv,chen2020dynamicconv,mildenhall2018kpn}. 
Recent AiO restoration methods combine these ideas by routing among low-rank branches or hypernetwork-generated bases, such as ABAIR~\citep{serrano2024abair},
LoRA-IR~\citep{ai2024lorair}, and HAIR~\citep{cao2024hair}. 
CEA is related to this compact adaptation family, but differs in both factor
generation and factor application: its low-rank factors are synthesized by
cross-attention queries that probe intermediate spatial features, and each
spatial token assembles its own residual update through dense signed affinities
over the generated rank-wise components. This makes CEA particularly suited to
spatially non-uniform and compositional degradations.

\noindent\textbf{Linear-attention-style computation.}
Efficient and linear attention methods reduce the quadratic cost of standard
attention by exploiting associative matrix products or kernelized feature maps
~\citep{shen2021efficient,katharopoulos2020transformers,choromanski2021performer}.
CEA does not propose a new general-purpose attention layer. 
Instead, we observe that the CEA residual has the same algebraic form as dot-product attention without softmax, \(QK^\top V\), where the keys and values are instance-conditioned routing bases and residual directions generated by the hyper-adapter. 
This view provides an efficient implementation of dense token-to-rank assembly.
\begin{figure}[h]
    \centering
    \includegraphics[width=1.0\textwidth]{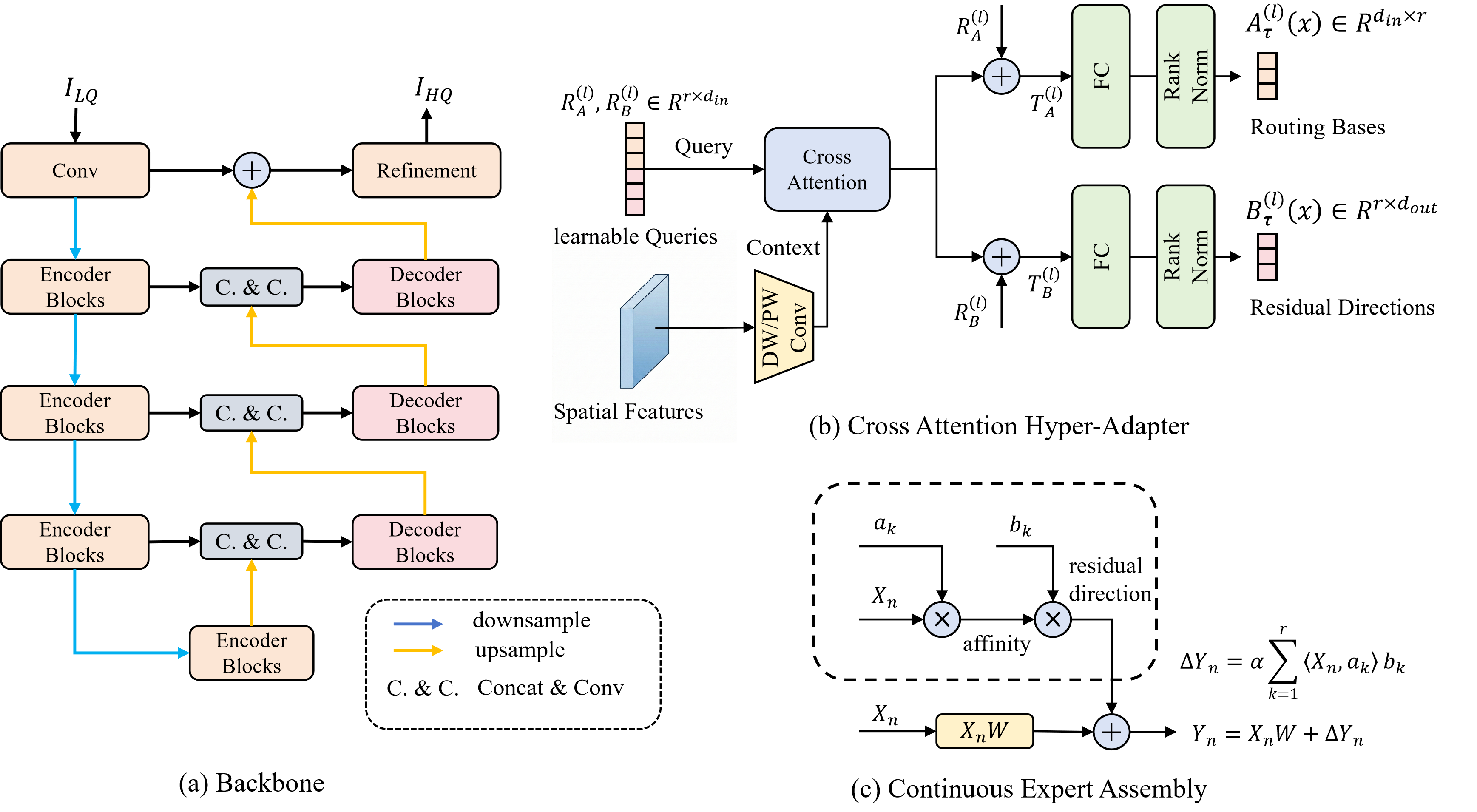}
    \caption{\textbf{Overall architecture of the proposed method.} (a) \textbf{Backbone.} We build upon a U-shaped Transformer backbone. To dynamically adapt to unknown degradations, our proposed modules are integrated into the decoder blocks to modulate the intermediate spatial features. 
    (b) \textbf{Cross-Attention Hyper-Adapter.} Spatial features are condensed via depthwise-pointwise (DW/PW) convolutions to provide local context. A compact set of learnable queries \((R_A^{(l)}, R_B^{(l)})\) probes this context via cross-attention. The contextualized queries are decoded into instance-conditioned routing bases \(A_{\tau}^{(l)}(x)\) and residual directions \(B_{\tau}^{(l)}(x)\), followed by RankNorm for stable assembly.
    (c) \textbf{Continuous Expert Assembly.} For each spatial token \(X_n\), CEA computes dense signed dot-product affinities \(\langle X_n,a_k\rangle\) with the generated routing bases. These affinities weight the corresponding residual directions \(b_k\) to assemble a token-specific residual update \(\Delta Y_n\), which is added to the base projection \(X_nW\) to produce the adapted feature \(Y_n\).
  }
    \label{fig:2} 
\end{figure}

% ------------------------------------------------
\section{Method}

\subsection{Problem Formulation \& The Mixture-of-Experts Baseline}
In All-in-One (AiO) image restoration, we are given a degraded observation $x \in \mathbb{R}^{H \times W \times 3}$ generated by an unknown degradation process $x = \mathcal{D}_{t}(y; \xi)$, where $t$ denotes a latent degradation type or composition and $\xi$ represents its severity and nuisance parameters. Our primary objective is to learn a unified model $F(\cdot)$ that reconstructs the clean image $\hat{y} = F(x) \approx y$ without prior knowledge of the explicit degradation parameters $(t, \xi)$.

To address the severe heterogeneity of diverse degradations within a single model, a representative strategy is \textbf{expert allocation} via conditional computation, such as (sparse) Mixture-of-Experts (MoE) and MoE-style expert branches explored in recent restoration models \citep{guo2024adaptir,zamfir2025moceir}. 
Formally, at decoder block $l$, we flatten the intermediate feature map of spatial size $H_l \times W_l$ into a token sequence $X^{(l)} \in \mathbb{R}^{N \times d_{\mathrm{in}}}$ with $N = H_l W_l$ spatial tokens, and use $X^{(l)}_n \in \mathbb{R}^{d_{\mathrm{in}}}$ to denote the $n$-th token (we omit the batch index for clarity).
For the $n$-th token $X^{(l)}_n$, a common \emph{MoE-style expert augmentation} adds expert-dependent residual branches on top of a shared projection $W \in \mathbb{R}^{d_{\mathrm{in}} \times d_{\mathrm{out}}}$, using a sparse router over a \emph{static} expert set $\{\mathcal{E}_i\}_{i=1}^E$:
\begin{equation}
Y^{(l)}_n = X^{(l)}_n W + \sum_{i=1}^{E} g_i(X^{(l)}_n) \mathcal{E}_i(X^{(l)}_n),
\end{equation}
where $g_i(\cdot)$ is a discrete routing mechanism (e.g., Top-$K$ Softmax) that assigns the token to specific experts.

\paragraph{Limitations in the restoration setting.} While sparse MoE is a natural way to allocate conditional capacity, it introduces additional design considerations for AiO restoration. First, Top-$K$ routing yields discontinuous expert assignments; in practice this can make optimization more sensitive and often requires capacity handling. Second, the expert parameters are typically shared across samples, so adaptation is limited to mixing a discrete set of functions even when degradations vary continuously in type and severity. These observations motivate a dense, continuous, and instance-conditioned alternative.

\subsection{Continuous Expert Assembly: Generated Low-Rank Routing}
\label{sec:cea_low_rank_routing}
\label{sec:attention_moe_view}

We propose \textbf{Continuous Expert Assembly} (CEA), which relaxes sparse
expert allocation into continuous low-rank residual assembly. 
Instead of selecting from a fixed pool of expert networks, CEA synthesizes rank-wise routing bases and residual directions from the current degraded image. 
Each spatial token then assembles an additive update over this generated subspace using dense signed inner products rather than sparse or probability-normalized routing weights.

With these design choices, we replace discrete expert allocation with a continuous, input-conditioned low-rank residual assembly rule. We use the term \emph{residual direction} for \(b^{(l)}_{\tau,k}(x)\) because it is an output-space vector that specifies the direction of the additive update contributed by the \(k\)-th rank component. It is not a standalone expert network. When needed, we refer to the pair \((a^{(l)}_{\tau,k}, b^{(l)}_{\tau,k})\) as a rank-wise component: \(a^{(l)}_{\tau,k}\) determines how strongly a token uses the component, while \(b^{(l)}_{\tau,k}\) specifies the direction added to the residual update. For a token $X^{(l)}_n$, the instance-conditioned continuous routing residual
\(\Delta Y^{(l)}_{\tau,n}\) for target projection \(\tau\) is defined as:

\begin{equation}
\Delta Y^{(l)}_{\tau,n} = \alpha \sum_{k=1}^{r} 
\underbrace{\langle X^{(l)}_n, a^{(l)}_{\tau,k}(x) \rangle}_{\substack{\text{signed} \\ \text{affinity}}}
\underbrace{b^{(l)}_{\tau,k}(x)}_{\substack{\text{residual}\\\text{direction}}},
\label{eq:moe_token}
\end{equation}
where $a^{(l)}_{\tau,k}(x) \in \mathbb{R}^{d_{\mathrm{in}}}$ is the $k$-th \emph{routing basis} (key) for target $\tau$, 
$b^{(l)}_{\tau,k}(x) \in \mathbb{R}^{d_{\mathrm{out}}}$ is the corresponding instance-conditioned \emph{residual direction}, and $\alpha = 1/r$ is a stabilizing scaling factor.

Crucially, when written for the full token sequence, Eq.~\eqref{eq:moe_token} has the same \emph{algebraic} form as dot-product attention without softmax, \(QK^\top V\). 
Since the effective keys and values correspond to the generated rank-$r$ factors, the update can be computed by two low-rank matrix multiplications, scaling linearly with the number of tokens \(N\) and the rank \(r\). 
Let $A^{(l)}_{\tau}(x) \in \mathbb{R}^{d_{\mathrm{in}} \times r}$ and $B^{(l)}_{\tau}(x) \in \mathbb{R}^{r \times d_{\mathrm{out}}}$ be the matrices formed by columns $a^{(l)}_{\tau,k}(x)$ and rows $b^{(l)}_{\tau,k}(x)$, respectively. The global parameter assembly becomes:

\begin{equation}
\begin{aligned}
G^{(l)}_{\tau}(x)
&\triangleq X^{(l)}A^{(l)}_{\tau}(x)
\in \mathbb{R}^{N\times r},\\
\Delta Y^{(l)}_{\tau}
&= \alpha\,G^{(l)}_{\tau}(x)B^{(l)}_{\tau}(x)\\
&= \alpha\,X^{(l)}A^{(l)}_{\tau}(x)B^{(l)}_{\tau}(x)\\
&= \alpha\,\mathrm{Attn}_{\mathrm{lin}}
\!\left(X^{(l)}, A^{(l)}_{\tau}(x)^\top, B^{(l)}_{\tau}(x)\right).
\end{aligned}
\label{eq:deltaY}
\end{equation}

Here \(\mathrm{Attn}_{\mathrm{lin}}(Q,K,V)=QK^\top V\) denotes dot-product
attention without softmax; the last line uses
\(Q=X^{(l)}\), \(K=A^{(l)}_{\tau}(x)^\top\), and
\(V=B^{(l)}_{\tau}(x)\). In this view, each row of
\(G^{(l)}_{\tau}(x)\) provides dense token-wise routing weights over the
\(r\) rank-wise components, and \(B^{(l)}_{\tau}(x)\) stores the corresponding
residual directions to be mixed.

Eq.~\eqref{eq:deltaY} gives an attention-style algebraic view of the CEA
residual. Unlike standard attention keys and values, however,
\(A^{(l)}_\tau(x)\) and \(B^{(l)}_\tau(x)\) are image-conditioned, shared across
tokens within the same layer and target, and produced by the hyper-adapter
rather than by projecting \(X^{(l)}\). CEA is therefore neither a uniform
hypernetwork-generated LoRA matrix nor soft expert mixing over a fixed bank.
Instead, token dependence arises from \(G^{(l)}_\tau(x)=X^{(l)}A^{(l)}_\tau(x)\), whose rows provide token-specific signed, softmax-free affinities over the generated rank-wise components; these affinities then combine the generated residual directions in \(B^{(l)}_\tau(x)\) to yield spatially varying residual updates.

\subsection{Cross-Attention Hyper-Adapter}
\label{sec:hyper_adapter}
The low-rank factors in CEA should be conditioned on the current degradation while remaining efficient to generate. A simple option is to pool the feature map into a global descriptor and predict \(A(x)\) and \(B(x)\) with an MLP. However, such global conditioning can discard spatial evidence that is important for localized and compositional degradations. We therefore use spatial query probing rather than global factor prediction.

To this end, we introduce a lightweight \textbf{Cross-Attention Hyper-Adapter} (illustrated in Figure~\ref{fig:2}). Instead of relying on a single global descriptor that can compress spatially heterogeneous degradation cues, we use a compact set of learnable queries as spatial probes. The queries interact with condensed spatial features via cross-attention to aggregate local degradation context for generating routing bases and residual directions.

Given the intermediate spatial features $X^{(l)}$ at the $l$-th block, we first condense their spatial resolution via depthwise-pointwise convolutions to yield $\hat{X}^{(l)}$. This operation efficiently reduces the computational overhead of the subsequent attention mechanism while encoding local structural patterns.  The learnable queries then interact with these condensed features via cross-attention to aggregate local degradation contexts:

\begin{equation}
T_{\phi}^{(l)}(x) = R_{\phi}^{(l)} + \mathrm{CrossAttn}\!\left(R_{\phi}^{(l)}, \hat{X}^{(l)}, \hat{X}^{(l)}\right), \quad \phi \in \{A, B\}.
\label{eq:T_phi}
\end{equation}

Finally, lightweight fully-connected (FC) heads decode these contextualized query representations into a set of \emph{routing bases} and \emph{residual directions} for continuous low-rank residual assembly.
We share $T_A^{(l)}(x)$ and $T_B^{(l)}(x)$ across targets, while using \emph{target-specific} heads to produce the corresponding routing bases and residual directions (i.e., matrices $A^{(l)}_{\tau}(x)$ and $B^{(l)}_{\tau}(x)$) for each modulated projection (e.g., $Q$ and $K$):
\begin{equation}
A^{(l)}_{\tau}(x) = \Big(\mathrm{FC}^{(l)}_{A,\tau}\big(T_A^{(l)}(x)\big)\Big)^{\top}, \qquad
B^{(l)}_{\tau}(x) = \mathrm{FC}^{(l)}_{B,\tau}\big(T_B^{(l)}(x)\big), \qquad \tau \in \{Q,K\}.
\label{eq:target_heads}
\end{equation}

\subsection{Stable Parameter Assembly and Injection}
\label{sec:ranknorm_gate}

\textbf{Resolving Scale Ambiguity (RankNorm).}
Dynamic factorization is inherently scale-ambiguous, as \(AB=(cA)(B/c)\).
To prevent norm drift and destabilization during training, we apply Rank-wise
Factor Normalization (RankNorm) to the generated factors. For each target
\(\tau\) and rank index \(k\),
\begin{equation}
\big(A^{(l)\prime}_{\tau}(x)\big)_{:,k}
=
\frac{\big(A^{(l)}_{\tau}(x)\big)_{:,k}}
{\|\big(A^{(l)}_{\tau}(x)\big)_{:,k}\|_2+\epsilon},
\qquad
\big(B^{(l)\prime}_{\tau}(x)\big)_{k,:}
=
\frac{\big(B^{(l)}_{\tau}(x)\big)_{k,:}}
{\|\big(B^{(l)}_{\tau}(x)\big)_{k,:}\|_2+\epsilon}.
\end{equation}

Crucially, this normalization is applied \emph{per instance} to prevent cross-sample coupling, ensuring that each dynamically synthesized rank component remains well-conditioned. In implementation, the residual is computed as two low-rank products,
\((X^{(l)}A^{(l)\prime}_\tau(x))B^{(l)\prime}_\tau(x)\), rather than forming and
applying a dense dynamic projection. This reduces the dominant computation
from a full \(N d_{\mathrm{in}} d_{\mathrm{out}}\) dynamic projection to
\(N d_{\mathrm{in}} r + N r d_{\mathrm{out}}\), with \(r \ll d_{\mathrm{in}},
d_{\mathrm{out}}\).

\textbf{Injection Targets.} The normalized factors are used to construct the residual $\Delta Y^{(l)}$. In our default architecture, we inject this dynamic residual into the Attention Query ($Q$) and Key ($K$) projections:
\begin{equation}
Q = X^{(l)} W_Q + \Delta Y_Q^{(l)}, \qquad K = X^{(l)} W_K + \Delta Y_K^{(l)}.
\end{equation}
In particular, $\Delta Y^{(l)}_{Q}$ and $\Delta Y^{(l)}_{K}$ are assembled by Eq.~\eqref{eq:deltaY} using their corresponding target-specific factors $(A^{(l)}_{Q}(x),B^{(l)}_{Q}(x))$ and $(A^{(l)}_{K}(x),B^{(l)}_{K}(x))$, respectively. By modulating \(Q\) and \(K\), CEA adjusts the backbone's attention patterns in a token-dependent manner, enabling spatially adaptive routing under the current degradation instance.

\paragraph{Training objective.}
We train the network with a reconstruction objective
\[
\mathcal{L}_{\mathrm{total}}
=
\mathcal{L}_{\mathrm{rec}}
+
\lambda_f \mathcal{L}_{\mathrm{fft}}.
\]
Here \(\mathcal{L}_{\mathrm{rec}}\) is the pixel-domain \(\ell_1\) loss.
Following recent frequency-aware restoration practice and the training recipe of
MoCE-IR \citep{zamfir2025moceir}, we also use a Fourier-domain reconstruction loss
\(\mathcal{L}_{\mathrm{fft}}\) as a shared regularizer.
This term is not part of the proposed CEA module; it is used for all matched
variants in the controlled analysis.

% ------------------------------------------------
\section{Experiments}
\subsection{Experiment Settings}
\textbf{Datasets and protocols.} We evaluate our Continuous Expert Assembly (CEA) framework following standard All-in-One (AiO) restoration protocols \citep{potlapalli2023promptir,zhang2023idr}. Specifically, we conduct experiments under the AIO-3 and AIO-5 settings, where a single model is trained on mixed degradations and tested on task-specific benchmarks, as well as under the CDD-11 protocol~\citep{guo2024onerestore} for compositional degradations involving combinations of up to three operators.

\textbf{Implementation details.}
We adopt an asymmetric U-shaped Transformer backbone following prior all-in-one restorers. By default, CEA modules are inserted into decoder blocks and modulate the attention \(Q/K\) projections. We optimize all CEA models and controlled variants using Adam~\citep{kingma2014adam} with a reconstruction objective consisting of a spatial \(\ell_1\) loss and a Fourier-domain reconstruction regularizer, following recent all-in-one restoration practice~\citep{zamfir2025moceir} and frequency-aware restoration objectives~\citep{jiang2021ffl}. This Fourier term is shared across matched variants and is not a label-based auxiliary loss. No degradation labels, prompts, or auxiliary classification/contrastive objectives are used.
% Unless otherwise stated, all main results are obtained without degradation labels,
% prompts, or auxiliary supervision during training.

\subsection{All-in-One Restoration on Standard AIO Benchmarks}
Table~\ref{tab:main_aio_summary} summarizes the average performance on AIO-3 and AIO-5; full per-task comparisons are provided in Appendix~\ref{app:full_aio3} and Appendix~\ref{app:full_aio5}.

\paragraph{AIO-3.}
In the lightweight regime, CEA-IR-S improves the average score from 32.57/0.916 to 32.89/0.918 compared with MoCE-IR-S while using fewer parameters. The gains are most visible on SOTS dehazing and Rain100L deraining, which involve spatially varying weather degradations. In the higher-capacity setting, CEA-IR improves the average score from 32.73/0.917 to 33.07/0.922. The denoising gains are smaller but consistent across noise levels, indicating that input-conditioned assembly does not sacrifice performance on more homogeneous degradations.

\paragraph{AIO-5.}
Compared with MoCE-IR-S, CEA-IR-S improves the average score from 30.08/0.913 to 30.80/0.919. The gain is mainly driven by deraining, deblurring, low-light enhancement, and a smaller improvement on denoising. On dehazing, CEA-IR-S obtains slightly lower PSNR than MoCE-IR-S but comparable SSIM, indicating that the average improvement does not rely on uniformly dominating every task. For heavy models, CEA-IR improves the average score from 30.58/0.919 to 31.13/0.922. The largest gains appear on dehazing and deraining, while the model remains competitive but slightly lower on GoPro deblurring compared with MoCE-IR.

\begin{table}[t]
\centering
\caption{\textbf{Main all-in-one restoration results.}
We report average PSNR/SSIM. Full per-task comparisons are provided in Appendix~\ref{app:full_aio3} and Appendix~\ref{app:full_aio5}.}
\label{tab:main_aio_summary}
\small
\begin{tabular}{llccccc}
\toprule
Benchmark & Scale & Method & Params. & Avg. PSNR & Avg. SSIM & Gain \\
\midrule
\multirow{2}{*}{AIO-3} & Light & MoCE-IR-S & 11M & 32.57 & .916 & -- \\
                     & Light & \textbf{CEA-IR-S} & \textbf{10M} & \textbf{32.89} & \textbf{.918} & \textbf{+0.32/+.002}
\\
\midrule
\multirow{2}{*}{AIO-3} & Heavy & MoCE-IR & 25M & 32.73 & .917 & -- \\
                     & Heavy & \textbf{CEA-IR} & \textbf{22M} & \textbf{33.07} & \textbf{.922} & \textbf{+0.34/+.005} \\
\midrule
\multirow{2}{*}{AIO-5} & Light & MoCE-IR-S & 11M & 30.08 & .913 & -- \\
                     & Light & \textbf{CEA-IR-S} & \textbf{10M} & \textbf{30.80} & \textbf{.919} & \textbf{+0.72/+.006}
\\
\midrule
\multirow{2}{*}{AIO-5} & Heavy & MoCE-IR & 25M & 30.58 & .919 & -- \\
                     & Heavy & \textbf{CEA-IR} & \textbf{22M} & \textbf{31.13} & \textbf{.922} & \textbf{+0.55/+.003} \\
\bottomrule
\end{tabular}
\end{table}

\subsection{Compositional Degradations on CDD-11}

Compositional degradations require generalization beyond individual degradation
operators to their interactions and spatial overlap. Table~\ref{tab:cdd11_group}
reports group-level results on the CDD-11 protocol~\citep{guo2024onerestore}.
Single, Double, and Triple denote category-wise averages over the corresponding
degradation groups, and Overall is the category-wise average over all 11
degradation categories following prior work.

\begin{table}[t]
\centering
\caption{\textbf{CDD-11 compositional restoration.}
We report group-level PSNR/SSIM on single, double, and triple degradation groups. Full per-category scores are provided in Appendix~\ref{app:full_cdd11}.}
\label{tab:cdd11_group}
\small
\begin{tabular}{lcccc}
\toprule
Method & Single & Double & Triple & Overall \\
\midrule
OneRestore & 31.68/.938 & 27.35/.866 & 24.84/.790 & 28.47/.878 \\
MoCE-IR-S & 32.54/.941 & 27.73/.869 & 25.40/.790 & 29.05/.881 \\
CEA-IR-S & \textbf{33.36/.943} & \textbf{28.51/.875} & \textbf{25.90/.798} & \textbf{29.79/.886} \\
\bottomrule
\end{tabular}
\end{table}

CEA-IR-S achieves the best overall average of 29.79/0.886 using 10M parameters,
improving over OneRestore and MoCE-IR-S by 1.32 dB and 0.74 dB, respectively.
The gains are consistent across single, double, and triple degradation groups,
with especially clear improvements on haze-related compositions in the full
per-category table. These results match the motivation of CEA: compositional
degradations often involve both global visibility changes and localized
degradation patterns, making it useful to generate an input-conditioned residual
subspace while allowing different spatial regions to assemble different updates.
We therefore analyze the roles of factor generation and spatial assembly
explicitly in Sec.~\ref{sec:controlled_analysis}.

\begin{table}[htbp]
\centering
\caption{\textbf{Computational demands.} FLOPs are computed on an input image
of size \(224 \times 224\). Runtime (mean \(\pm\) std in milliseconds) is re-evaluated on a single local NVIDIA RTX 4090 GPU using officially released
codebases under identical input configurations.}
\label{tab:flops}
% \resizebox{\linewidth}{!}{
\begin{tabular}{lcccc}
\toprule
Method & Params. & Memory & FLOPs & Runtime (ms) \\
\midrule
AirNet  & 8.93M & 4829M & 238G & 42.13 $\pm$ 0.13 \\
PromptIR  & 35.59M & 9830M & 132G & 40.86 $\pm$ 0.24 \\
MoCE-IR  & 25.35M & 5887M & 80.59G & 43.10 $\pm$ 0.27 \\
MoCE-IR-S  & 11.47M & 4228M & 36.93G & 41.63 $\pm$ 0.41 \\
CEA-IR-S (Ours) & 9.71M & \textbf{3517M} & \textbf{28.03G} & \textbf{29.92 $\pm$ 0.25} \\
\bottomrule
\end{tabular}
% }
\end{table}

\subsection{Qualitative Evaluation}

Representative qualitative comparisons are provided in Appendix~\ref{app:qualitative_results}. Across individual and compositional degradations, CEA reduces localized residual artifacts such as rain streaks, haze boundaries, and low-light texture distortions, while preserving fine structures. These visual trends are consistent with the quantitative gains in Tables~\ref{tab:main_aio_summary}--\ref{tab:cdd11_group}.

\subsection{Efficiency}
\label{sec:efficiency}
Table~\ref{tab:flops} reports parameter count, memory, FLOPs, and runtime measured under the same input setting. CEA-IR-S uses 9.71M parameters and 28.03 GFLOPs, both lower than MoCE-IR-S and PromptIR, and runs in 29.92 ms on an RTX 4090. The efficiency mainly comes from applying the image-conditioned update through
two low-rank matrix multiplications, rather than executing sparse expert
branches or applying an equivalent dense dynamic projection. This yields a
favorable accuracy--efficiency trade-off among the compared all-in-one
restorers.

% ---------------------------

\subsection{Controlled Analysis}
\label{sec:controlled_analysis}

\paragraph{Disentangling factor source and routing rule.}
Table~\ref{tab:factor_routing_ablation} tests whether the gain comes from the
specific mechanism of CEA rather than from simply adding low-rank residual
capacity. Static variants replace the input-conditioned hyper-adapter with
per-block learnable factors \(A,B\) that are independent of the input image.
Dynamic variants generate \(A(x),B(x)\) with the CEA hyper-adapter. Top-2
variants replace CEA's dense assembly with sparse softmax routing over the top
two rank components, whereas Dense uses CEA's softmax-free signed dot-product
assembly in Eq.~\eqref{eq:deltaY}. All variants share the same lightweight
backbone, Q/K injection, rank \(r=8\), and reconstruction losses.

\begin{table}[t]
\centering
\caption{
\textbf{Factorial ablation on CDD-11.}
All variants use the same lightweight backbone, Q/K injection, rank \(r=8\), and reconstruction losses. ``Static'' learns input-independent low-rank factors, while ``Dynamic'' generates factors with the CEA hyper-adapter. ``Top-2'' applies sparse softmax routing over the top two rank components; ``Dense'' uses CEA's softmax-free signed dot-product assembly.
}
\label{tab:factor_routing_ablation}
\small
\begin{tabular}{llcccc}
\toprule
Factor source & Routing rule & Single & Double & Triple & Avg. \\
\midrule
Static & Top-2 softmax & 32.15/.939 & 27.50/.867 & 25.16/.787 & 28.77/.879 \\
Static & Dense signed dot-product & 32.35/.940 & 27.77/.869 & 25.40/.790 & 29.01/.881 \\
Dynamic & Top-2 softmax & 32.50/.941 & 27.94/.870 & 25.76/.792 & 29.20/.882 \\
Dynamic & Dense signed dot-product, CEA & \textbf{33.36/.943} & \textbf{28.51/.875} & \textbf{25.90/.798} & \textbf{29.79/.886} \\
\bottomrule
\end{tabular}
\end{table}

The factorial ablation shows that the gain does not come from merely adding
low-rank residual capacity. With Top-2 routing, dynamic factors improve the
overall average from 28.77/0.879 to 29.20/0.882. With dense assembly, the same
change improves the overall average from 29.01/0.881 to 29.79/0.886, showing
the importance of input-conditioned factor generation. Static dense assembly remains close to the backbone-only result in Table~\ref{tab:app_ablation_inject}, suggesting that input-independent low-rank factors are not sufficient for compositional restoration. Dense signed assembly also improves over Top-2 routing under both factor sources, and the gain is larger with dynamic factors (+0.59 dB) than with static factors (+0.24 dB). This interaction supports the central design of CEA: rank-wise components should be generated from the current image and used through dense token-wise assembly. Additional controlled studies on spatial query probing, RankNorm, rank sensitivity, and injection targets are provided in Appendix~\ref{app:additional_ablations}.

\section{Limitations}
This work focuses on standard individual and compositional restoration
benchmarks. Evaluating CEA on additional restoration settings would further
clarify its generalization behavior. The current implementation uses a U-shaped
Transformer backbone and modulates attention projections; extending CEA to
other backbones and dense prediction tasks is left for future work. As with
other image enhancement methods, improved restoration may raise privacy concerns
or unintended downstream use, so responsible use should consider consent and
appropriate data handling.

\section{Conclusion}
We proposed \emph{Continuous Expert Assembly} (CEA), a parameter-efficient framework for all-in-one image restoration under unknown and compositional degradations. CEA changes expert adaptation from selecting among fixed experts to assembling token-specific residual updates from input-generated rank-wise components. A lightweight Cross-Attention Hyper-Adapter probes intermediate spatial features to synthesize routing bases and residual directions, while dense signed affinities assemble spatially adaptive residuals without discrete expert selection. Experiments on AIO-3, AIO-5, and CDD-11 show improved average restoration quality, especially on compositional degradations, while maintaining favorable parameters, FLOPs, and runtime efficiency.

\bibliographystyle{plainnat}
\bibliography{main}

@String(CVPR  = {IEEE Conf. Comput. Vis. Pattern Recog.})

@String(ICCV  = {Int. Conf. Comput. Vis.})

@String(ECCV  = {Eur. Conf. Comput. Vis.})

@String(NeurIPS = {Adv. Neural Inform. Process. Syst.})

@String(ICML  = {Int. Conf. Mach. Learn.})

@String(ICLR  = {Int. Conf. Learn. Represent.})

@String(CVPRW = {IEEE Conf. Comput. Vis. Pattern Recog. Worksh.})

@String(AAAI  = {AAAI})

@String(CVPR  = {CVPR})

@String(ICCV  = {ICCV})

@String(ECCV  = {ECCV})

@String(NeurIPS = {NeurIPS})

@String(ICML  = {ICML})

@String(ICLR  = {ICLR})

@String(CVPRW = {CVPRW})

@inproceedings{liang2021swinir,
  title     = {SwinIR: Image Restoration Using Swin Transformer},
  author    = {Jingyun Liang and Jiezhang Cao and Guolei Sun and Kai Zhang and Luc Van Gool and Radu Timofte},
  booktitle = {2021 IEEE/CVF International Conference on Computer Vision Workshops (ICCVW)},
  pages     = {1833--1844},
  year      = {2021},
  doi       = {10.1109/ICCVW54120.2021.00210}
}

@inproceedings{zamir2022restormer,
  title     = {Restormer: Efficient Transformer for High-Resolution Image Restoration},
  author    = {Syed Waqas Zamir and Aditya Arora and Salman Khan and Munawar Hayat and Fahad Shahbaz Khan and Ming-Hsuan Yang},
  booktitle = {2022 IEEE/CVF Conference on Computer Vision and Pattern Recognition (CVPR)},
  pages     = {5728--5739},
  year      = {2022},
  doi       = {10.1109/CVPR52688.2022.00564}
}

@inproceedings{wang2022uformer,
  title     = {Uformer: A General U-Shaped Transformer for Image Restoration},
  author    = {Zhendong Wang and Xiaodong Cun and Jianmin Bao and Wengang Zhou and Jianzhuang Liu and Houqiang Li},
  booktitle = {2022 IEEE/CVF Conference on Computer Vision and Pattern Recognition (CVPR)},
  month     = {June},
  pages     = {17662--17672},
  year      = {2022},
  doi       = {10.1109/CVPR52688.2022.01716}
}

@inproceedings{chen2022nafnet,
  title     = {Simple Baselines for Image Restoration},
  author    = {Liangyu Chen and Xiaojie Chu and Xiangyu Zhang and Jian Sun},
  booktitle = {Computer Vision -- ECCV 2022},
  pages     = {17--33},
  year      = {2022},
  doi       = {10.1007/978-3-031-20071-7\_2}
}

@inproceedings{chen2022cat,
  title     = {Cross Aggregation Transformer for Image Restoration},
  author    = {Zheng Chen and Yulun Zhang and Jinjin Gu and Yongbing Zhang and Linghe Kong and Xin Yuan},
  booktitle = {Advances in Neural Information Processing Systems (NeurIPS)},
  volume    = {35},
  pages     = {25478--25490},
  year      = {2022}
}

@inproceedings{li2022airnet,
  title     = {All-In-One Image Restoration for Unknown Corruption},
  author    = {Boyun Li and Xiao Liu and Peng Hu and Zhongqin Wu and Jiancheng Lv and Xi Peng},
  booktitle = {2022 IEEE/CVF Conference on Computer Vision and Pattern Recognition (CVPR)},
  pages     = {17431--17441},
  year      = {2022},
  doi       = {10.1109/CVPR52688.2022.01693}
}

@inproceedings{potlapalli2023promptir,
  title     = {PromptIR: prompting for all-in-one blind image restoration},
  author    = {Vaishnav Potlapalli and Syed Waqas Zamir and Salman H. Khan and Fahad Shahbaz Khan},
  booktitle = {Advances in Neural Information Processing Systems (NeurIPS)},
  year      = {2023},
  pages     = {71275--71293},
  doi       = {10.52202/075280-3121}
}

@INPROCEEDINGS{zamfir2025moceir,
  author={Zamfir, Eduard and Wu, Zongwei and Mehta, Nancy and Tan, Yuedong and Paudel, Danda Pani and Zhang, Yulun and Timofte, Radu},
  booktitle={2025 IEEE/CVF Conference on Computer Vision and Pattern Recognition (CVPR)}, 
  title={Complexity Experts are Task-Discriminative Learners for Any Image Restoration}, 
  year={2025},
  pages={12753-12763},
  doi={10.1109/CVPR52734.2025.01190}}

@inproceedings{guo2024onerestore,
  title     = {OneRestore: A Universal Restoration Framework for Composite Degradation},
  author    = {Yu Guo and Yuan Gao and Yuxu Lu and Huilin Zhu and Ryan Wen Liu and Shengfeng He},
  booktitle = {Computer Vision -- ECCV 2024},
  pages     = {255--272},
  year      = {2025},
  doi       = {10.1007/978-3-031-72655-2\_15}
}

@inproceedings{hu2022lora,
  author    = {Edward J. Hu and Yelong Shen and Phillip Wallis and Zeyuan Allen-Zhu and Yuanzhi Li and Shean Wang and Lu Wang and Weizhu Chen},
  title     = {LoRA: Low-Rank Adaptation of Large Language Models},
  booktitle = {International Conference on Learning Representations (ICLR)},
  year      = {2022},
  doi       = {10.48550/arXiv.2106.09685}
}

@inproceedings{liu2024dora,
  author    = {Liu, Shih-Yang and Wang, Chien-Yi and Yin, Hongxu and Molchanov, Pavlo and Wang, Yu-Chiang Frank and Cheng, Kwang-Ting and Chen, Min-Hung},
  title     = {{D}o{RA}: Weight-Decomposed Low-Rank Adaptation},
  booktitle = {Proceedings of the 41st International Conference on Machine Learning (ICML)},
  year      = {2024},
  volume    = {235},
  pages     = {32100--32121},
  doi      = {10.48550/arXiv.2402.09353}  
}

@inproceedings{guo2024adaptir,
  title     = {Parameter Efficient Adaptation for Image Restoration with Heterogeneous Mixture-of-Experts},
  author    = {Guo, Hang and Dai, Tao and Bai, Yuanchao and Chen, Bin and Ren, Xudong and Zhu, Zexuan and Xia, Shu-Tao},
  booktitle = {Advances in Neural Information Processing Systems (NeurIPS)},
  year      = {2024},
  pages     = {13522--13547},
  doi       = {10.52202/079017-0432}
}

@article{kingma2014adam,
  title   = {Adam: A Method for Stochastic Optimization},
  author  = {Diederik P. Kingma and Jimmy Ba},
  journal = {arXiv preprint arXiv:1412.6980},
  year    = {2014},
  doi     = {10.48550/arXiv.1412.6980}
}

@article{cao2024hair,
  title   = {HAIR: Hypernetworks-based All-in-One Image Restoration},
  author  = {Jin Cao and Yi Cao and Li Pang and Deyu Meng and Xiangyong Cao},
  journal = {arXiv preprint arXiv:2408.08091},
  year    = {2024},
  doi     = {10.48550/arXiv.2408.08091}
}

@inproceedings{conde2024instructir,
  title     = {InstructIR: High-Quality Image Restoration Following Human Instructions},
  author    = {Conde, Marcos V. and Geigle, Gregor and Timofte, Radu},
  booktitle = {Computer Vision -- ECCV 2024},
  year      = {2024},
  pages = {1--21},
  doi       = {10.1007/978-3-031-72764-1\_1}
}

@inproceedings{houlsby2019adapters,
  title     = {Parameter-Efficient Transfer Learning for {NLP}},
  author    = {Houlsby, Neil and Giurgiu, Andrei and Jastrzebski, Stanislaw and Morrone, Bruna and De Laroussilhe, Quentin and Gesmundo, Andrea and Attariyan, Mona and Gelly, Sylvain},
  booktitle = {Proceedings of the 36th International Conference on Machine Learning (ICML)},
  year      = {2019},
  pages = 	 {2790--2799},
  volume = 	 {97}

}

@inproceedings{zhang2023adalora,
  title     = {Adaptive Budget Allocation for Parameter-Efficient Fine-Tuning},
  author    = {Zhang, Qingru and Chen, Minshuo and Bukharin, Alexander and
               Karampatziakis, Nikos and He, Pengcheng and Cheng, Yu and
               Chen, Weizhu and Zhao, Tuo},
  booktitle = {International Conference on Learning Representations (ICLR)},
  year      = {2023},
  note       = {10.48550/arXiv.2303.10512}
}

@inproceedings{ha2016hypernetworks,
  title     = {HyperNetworks},
  author    = {Ha, David and Dai, Andrew and Le, Quoc V.},
  booktitle = {International Conference on Learning Representations (ICLR)},
  year      = {2017}
}

@article{li2023pip,
  title         = {Prompt-In-Prompt Learning for Universal Image Restoration},
  author        = {Li, Zilong and Lei, Yiming and Ma, Chenglong and Zhang, Junping and Shan, Hongming},
  year          = {2023},
 journal = {arXiv preprint arXiv:2312.05038},
  doi           = {10.48550/arXiv.2312.05038},
}

@inproceedings{jia2016dynamic,
  title     = {Dynamic filter networks},
  author    = {De Brabandere, Bert and Jia, Xu and Tuytelaars, Tinne and Van Gool, Luc},
  booktitle = {Advances in Neural Information Processing Systems (NeurIPS)},
  pages = {667–675},
  year      = {2016},
  note = {10.48550/arXiv.1605.09673}
}

@inproceedings{yang2019condconv,
  title     = {CondConv: conditionally parameterized convolutions for efficient inference},
  author    = {Yang, Brandon and Bender, Gabriel and Le, Quoc V. and Ngiam, Jiquan},
  booktitle = {Advances in Neural Information Processing Systems (NeurIPS)},
  editor    = {Wallach, Hanna M. and Larochelle, Hugo and Beygelzimer, Alina and d'Alch{\'e}-Buc, Florence and Fox, Emily B. and Garnett, Roman},
  pages     = {1307--1318},
  year      = {2019}
}

@inproceedings{shazeer2017moe,
  title     = {Outrageously Large Neural Networks: The Sparsely-Gated Mixture-of-Experts Layer},
  author    = {Shazeer, Noam and Mirhoseini, Azalia and Maziarz, Krzysztof and Davis, Andy and Le, Quoc V. and Hinton, Geoffrey E. and Dean, Jeff},
  booktitle = {International Conference on Learning Representations (ICLR)},
  year      = {2017},
  note       = {10.48550/arXiv.1701.06538}
}

@article{fedus2022switch,
  title   = {Switch Transformers: Scaling to Trillion Parameter Models with Simple and Efficient Sparsity},
  author  = {Fedus, William and Zoph, Barret and Shazeer, Noam},
  journal = {Journal of Machine Learning Research},
  year    = {2022},
  volume  = {23},
  number  = {120},
  pages   = {1--39}
}

@inproceedings{riquelme2021vmoe,
  title     = {Scaling vision with sparse mixture of experts},
  author    = {Riquelme, Carlos and Puigcerver, Joan and Mustafa, Basil and Neumann, Maxim and Jenatton, Rodolphe and Pinto, Andr\'{e} Susano and Keysers, Daniel and Houlsby, Neil},
  booktitle = {Advances in Neural Information Processing Systems (NeurIPS)},
  year      = {2021},
  pages     = {8583--8595},
  note       = {10.48550/arXiv.2106.05974}
}

@inproceedings{jiang2021ffl,
  title     = {Focal Frequency Loss for Image Reconstruction and Synthesis},
  author    = {Jiang, Liming and Dai, Bo and Wu, Wayne and Loy, Chen Change},
  booktitle = {2021 IEEE/CVF International Conference on Computer Vision (ICCV)},
  month     = {October},
  year      = {2021},
  pages     = {13899--13909},
  doi       = {10.1109/ICCV48922.2021.01366}
}

@inproceedings{zhang2023idr,
  title     = {Ingredient-oriented Multi-Degradation Learning for Image Restoration},
  author    = {Zhang, Jinghao and Huang, Jie and Yao, Mingde and Yang, Zizheng and Yu, Hu and Zhou, Man and Zhao, Feng},
  booktitle = {2023 IEEE/CVF Conference on Computer Vision and Pattern Recognition (CVPR)},
  year      = {2023},
  pages     = {5825--5835},
  publisher = {IEEE},
  doi       = {10.1109/CVPR52729.2023.00564}
}

@article{zhang2017dncnn,
  title   = {Beyond a Gaussian Denoiser: Residual Learning of Deep CNN for Image Denoising},
  author  = {Zhang, Kai and Zuo, Wangmeng and Chen, Yunjin and Meng, Deyu and Zhang, Lei},
  journal = {IEEE Transactions on Image Processing},
  volume  = {26},
  number  = {7},
  pages   = {3142--3155},
  year    = {2017},
  doi     = {10.1109/TIP.2017.2662206}
}

@inproceedings{chen2021ipt,
  title     = {Pre-Trained Image Processing Transformer},
  author    = {Chen, Hanting and Wang, Yunhe and Guo, Tianyu and Xu, Chang and Deng, Yiping and Liu, Zhenhua and Ma, Siwei and Xu, Chunjing and Xu, Chao and Gao, Wen},
  booktitle = {2021 IEEE/CVF Conference on Computer Vision and Pattern Recognition (CVPR)},
  year      = {2021},
  pages     = {12294--12305},
  doi       = {10.1109/CVPR46437.2021.01212}
}

@inproceedings{tu2022maxim,
  title     = {{MAXIM}: Multi-Axis {MLP} for Image Processing},
  author    = {Tu, Zhengzhong and Talebi, Hossein and Zhang, Han and Yang, Feng and Milanfar, Peyman and Bovik, Alan and Li, Yinxiao},
  booktitle = {2022 IEEE/CVF Conference on Computer Vision and Pattern Recognition (CVPR)},
  year      = {2022},
  pages     = {5759--5770},
  doi       = {10.1109/CVPR52688.2022.00568}
 
}

@inproceedings{liu2022tape,
  title     = {{TAPE}: Task-Agnostic Prior Embedding for Image Restoration},
  author    = {Liu, Lin and Xie, Lingxi and Zhang, Xiaopeng and Yuan, Shanxin and Chen, Xiangyu and Zhou, Wengang and Li, Houqiang and Tian, Qi},
  booktitle = {Computer Vision -- ECCV 2022},
  year      = {2022},
  doi       = {10.1007/978-3-031-19797-0\_26}
}

@inproceedings{zamir2021mprnet,
  author    = {Zamir, Syed Waqas and Arora, Aditya and Khan, Salman and Hayat, Munawar and Khan, Fahad Shahbaz and Yang, Ming-Hsuan and Shao, Ling},
  title     = {Multi-Stage Progressive Image Restoration},
  booktitle = {2021 IEEE/CVF Conference on Computer Vision and Pattern Recognition (CVPR)},
  year      = {2021},
  pages     = {14816-14826},
  doi = {10.1109/CVPR46437.2021.01458}
}

@inproceedings{chen2021hinet,
  author    = {Chen, Liangyu and Lu, Xin and Zhang, Jie and Chu, Xiaojie and Chen, Chengpeng},
  title     = {HINet: Half Instance Normalization Network for Image Restoration},
  booktitle = {2021 IEEE/CVF Conference on Computer Vision and Pattern Recognition Workshops (CVPRW)},
  month     = {June},
  year      = {2021},
  pages     = {182-192},
  doi = {10.1109/CVPRW53098.2021.00027}
}

@inproceedings{valanarasu2022transweather,
  author    = {Jose Valanarasu, Jeya Maria and Yasarla, Rajeev and Patel, Vishal M.},
  title     = {TransWeather: Transformer-based Restoration of Images Degraded by Adverse Weather Conditions},
  booktitle = {2022 IEEE/CVF Conference on Computer Vision and Pattern Recognition (CVPR)},
  year      = {2022},
  pages     = {2343--2353},
  doi = {10.1109/CVPR52688.2022.00239}
}

@article{ma2023prores,
  title        = {ProRes: Exploring Degradation-aware Visual Prompt for Universal Image Restoration},
  author       = {Jiaqi Ma and Tianheng Cheng and Guoli Wang and Qian Zhang and Xinggang Wang and Lefei Zhang},
  journal      = {arXiv preprint arXiv:2306.13653},
  year         = {2023},
  doi          = {10.48550/ARXIV.2306.13653}
}

@inproceedings{lepikhin2020gshard,
  author    = {Lepikhin, Dmitry and Lee, HyoukJoong and Xu, Yuanzhong and Chen, Dehao and Firat, Orhan and Huang, Yanping and Krikun, Maxim and Shazeer, Noam and Chen, Zhifeng},
  title     = {GShard: Scaling Giant Models with Conditional Computation and Automatic Sharding},
  booktitle = {International Conference on Learning Representations (ICLR)},
  year      = {2021}
}

@article{Fan2021DecoupledLearningPIO,
  title   = {A General Decoupled Learning Framework for Parameterized Image Operators},
  author  = {Fan, Qingnan and Chen, Dongdong and Yuan, Lu and Hua, Gang and Yu, Nenghai and Chen, Baoquan},
  journal = {IEEE Transactions on Pattern Analysis and Machine Intelligence},
  year    = {2021},
  volume={43},  
  pages={33-47},
  doi = {10.1109/TPAMI.2019.2925793}
}

@inproceedings{Mou2022DGUNet,
  title     = {Deep Generalized Unfolding Networks for Image Restoration},
  author    = {Mou, Chong and Wang, Qian and Zhang, Jian},
  booktitle = {2022 IEEE/CVF Conference on Computer Vision and Pattern Recognition (CVPR)},
  year      = {2022},
  doi = {10.1109/CVPR52688.2022.01688}
}

@inproceedings{Guo2024MambaIR,
  title     = {MambaIR: A Simple Baseline for Image Restoration with State-Space Model},
  author    = {Guo, Hang and Li, Jinmin and Dai, Tao and Ouyang, Zhihao and Ren, Xudong and Xia, Shu-Tao},
  booktitle = {Computer Vision -- ECCV 2024},
  year      = {2024},
  doi = {10.1007/978-3-031-72649-1\_13}
}

@article{Wang2023GridFormer,
  title   = {GridFormer: Residual Dense Transformer with Grid Structure for Image Restoration in Adverse Weather Conditions},
  author  = {Wang, Tao and Zhang, Kaihao and Shao, Ziqian and Luo, Wenhan and Stenger, Bjorn and Lu, Tong and Kim, Tae-Kyun and Liu, Wei and Li, Hongdong},
  journal = {International Journal of Computer Vision},
  year    = {2024},    
  volume  = {132},
  pages   = {4541--4563},
  doi  = {10.1007/s11263-024-02056-0}
}

@inproceedings{puigcerver2024softmoe,
  author    = {Puigcerver, Joan and Riquelme Ruiz, Carlos and Mustafa, Basil and Houlsby, Neil},
  title     = {From Sparse to Soft Mixtures of Experts},
  booktitle = {International Conference on Learning Representations},
   pages = {28435--28445},
  year      = {2024}

}

@inproceedings{choromanski2021performer,
  author    = {Krzysztof Choromanski and Valerii Likhosherstov and David Dohan and Xingyou Song and Andreea Gane and Tam{\'{a}}s Sarl{\'{o}}s and Peter Hawkins and Jared Davis and Afroz Mohiuddin and Lukasz Kaiser and David Belanger and Lucy J. Colwell and Adrian Weller},
  title     = {Rethinking Attention with Performers},
  booktitle = {International Conference on Learning Representations (ICLR)},
  year      = {2021}
}

@inproceedings{mildenhall2018kpn,
  author    = {Mildenhall, Ben and Barron, Jonathan T. and Chen, Jiawen and Sharlet, Dillon and Ng, Ren and Carroll, Robert},
  title     = {Burst Denoising with Kernel Prediction Networks},
  booktitle = {2018 IEEE/CVF Conference on Computer Vision and Pattern Recognition (CVPR)},
  year      = {2018},
  pages     = {2502--2510},
  doi       = {10.1109/CVPR.2018.00265}
}

@inproceedings{chen2020dynamicconv,
  author    = {Chen, Yinpeng and Dai, Xiyang and Liu, Mengchen and Chen, Dongdong and Yuan, Lu and Liu, Zicheng},
  title     = {Dynamic Convolution: Attention Over Convolution Kernels},
  booktitle = {2020 IEEE/CVF Conference on Computer Vision and Pattern Recognition (CVPR)},
  year      = {2020},
  pages     = {11027--11036},
  doi       = {10.1109/CVPR42600.2020.01104}
}

@inproceedings{zhu2023wgws,
  author       = {Zhu, Yurui and Wang, Tianyu and Fu, Xueyang and Yang, Xuanyu and Guo, Xin and Dai, Jifeng and Qiao, Yu and Hu, Xiaowei},
  title        = {Learning Weather-General and Weather-Specific Features for Image Restoration Under Multiple Adverse Weather Conditions},
  booktitle    = {2023 IEEE/CVF Conference on Computer Vision and Pattern Recognition (CVPR)},
  pages={21747-21758},
  year         = {2023},
  doi          = {10.1109/CVPR52729.2023.02083}
}

@article{ozdenizci2023weatherdiffusion,
  author       = {Ozan {\"{O}}zdenizci and Robert Legenstein},
  title        = {Restoring Vision in Adverse Weather Conditions With Patch-Based Denoising Diffusion Models},
  journal      = {IEEE Transactions on Pattern Analysis and Machine Intelligence},
  volume       = {45},
  number       = {8},
  pages={10346--10357},
  year         = {2023},
  doi          = {10.1109/TPAMI.2023.3238179}
}

@article{serrano2024abair,
  title         = {Adaptive Blind All-in-One Image Restoration},
  author        = {David Serrano-Lozano and Luis Herranz and Shaolin Su and Javier Vazquez-Corral},
  year          = {2025},
  journal      = {arXiv preprint arXiv:2411.18412},
  doi           = {10.48550/arXiv.2411.18412}
}

@article{ai2024lorair,
  title         = {LoRA-IR: Taming Low-Rank Experts for Efficient All-in-One Image Restoration},
  author        = {Yuang Ai and Huaibo Huang and Ran He},
  year          = {2024},
  journal       = {arXiv preprint arXiv:2410.15385},
  doi           = {10.48550/arXiv.2410.15385}
}

@article{tian2020brdnet,
  title     = {Image denoising using deep {CNN} with batch renormalization},
  author    = {Chunwei Tian and Yong Xu and Wangmeng Zuo},
  journal   = {Neural Networks},
volume = {121},
pages = {461-473},
  year      = {2020},
  doi       = {10.1016/j.neunet.2019.08.022}
}

@inproceedings{gao2019dynamicdeblur,
  title     = {Dynamic Scene Deblurring With Parameter Selective Sharing and Nested Skip Connections},
  author    = {Gao, Hongyun and Tao, Xin and Shen, Xiaoyong and Jia, Jiaya},
  booktitle = {2019 IEEE/CVF Conference on Computer Vision and Pattern Recognition (CVPR)},
  year      = {2019},
  pages={3843-3851},
  url       = {10.1109/CVPR.2019.00397}
}

@inproceedings{dong2020fdgan,
  title     = {{FD-GAN}: Generative Adversarial Networks with Fusion-Discriminator for Single Image Dehazing},
  author    = {Dong, Yu and Liu, Yihao and Zhang, He and Chen, Shifeng and Qiao, Yu},
  booktitle = { Proceedings of the AAAI Conference on Artificial Intelligence},
  year      = {2020},
  doi       = {10.1609/aaai.v34i07.6701}
}

@inproceedings{duan2024uniprocessor,
  title     = {UniProcessor: A Text-Induced Unified Low-Level Image Processor},
  author    = {Duan, Huiyu and Min, Xiongkuo and Wu, Sijing and Shen, Wei and Zhai, Guangtao},
  booktitle = {Computer Vision -- ECCV 2024},
  pages     = {180--199},
  year      = {2024},
  doi       = {10.1007/978-3-031-72855-6\_11}
}

@inproceedings{luo2024daclip,
  title     = {Controlling Vision-Language Models for Multi-Task Image Restoration},
  author    = {Luo, Ziwei and Gustafsson, Fredrik K. and Zhao, Zheng and Sj{\"o}lund, Jens and Sch{\"o}n, Thomas B.},
  booktitle = {International Conference on Learning Representations (ICLR)},
  year      = {2024}
}

@inproceedings{Zeng2025vlu,
  title     = {Vision-Language Gradient Descent-driven All-in-One Deep Unfolding Networks},
  author    = {Zeng, Haijin and Wang, Xiangming and Chen, Yongyong and Su, Jingyong and Liu, Jie},
  booktitle = {2025 IEEE/CVF Conference on Computer Vision and Pattern Recognition (CVPR)},
  year      = {2025},
  doi = {10.1109/CVPR52734.2025.00705}
}

@inproceedings{Wu2025art,
  title     = {Harmony in Diversity: Improving All-in-One Image Restoration via Multi-Task Collaboration},
  author    = {Wu, Gang and Jiang, Junjun and Jiang, Kui and Liu, Xianming},
  booktitle = {Proceedings of the 32nd ACM International Conference on Multimedia},
  year      = {2024},
  doi = {10.1145/3664647.3680762}
}

@inproceedings{shen2021efficient,
  title     = {Efficient Attention: Attention with Linear Complexities},
  author    = {Shen, Zhuoran and Zhang, Mingyuan and Zhao, Haiyu and Yi, Shuai and Li, Hongsheng},
  booktitle = {2021 IEEE Winter Conference on Applications of Computer Vision (WACV)},
  pages     = {3530-3538},
  year      = {2021},
  doi       = {10.1109/WACV48630.2021.00357}
}

@inproceedings{katharopoulos2020transformers,
  title     = {Transformers are {RNN}s: Fast Autoregressive Transformers with Linear Attention},
  author    = {Katharopoulos, Angelos and Vyas, Apoorv and Pappas, Nikolaos and Fleuret, Fran{\c{c}}ois},
  booktitle = {Proceedings of the 37th International Conference on Machine Learning},
  series    = {Proceedings of Machine Learning Research},
  volume    = {119},
  pages     = {5156--5165},
  year      = {2020}
}

@article{ren2026efficient,
  title   = {Efficient Degradation-agnostic Image Restoration via Channel-Wise Functional Decomposition and Manifold Regularization},
  author  = {Bin Ren and Yawei Li and Xu Zheng and Yuqian Fu and Danda Pani Paudel and Hong Liu and Ming-Hsuan Yang and Luc Van Gool and Nicu Sebe},
  journal = {arXiv preprint arXiv:2505.18679},
  year    = {2026},
  doi     = {10.48550/arXiv.2505.18679}
}

%%%%%%%%%%%%%%%%%%%%%%%%%%%%%%%%%%%%%%%%%%%%%%%%%%%%%%%%%%%%

\appendix

\section{Additional Method Details}
\label{app:method_details}

\subsection{Cross-Attention Hyper-Adapter Details}
\label{app:hyper_adapter_details}

\paragraph{Feature condensation.}
For each CEA-equipped decoder block, the hyper-adapter first condenses the
intermediate feature map before cross-attention. The condensation module
consists of a depthwise \(3\times3\) convolution followed by a pointwise
\(1\times1\) convolution. The depthwise convolution uses stride \(s_c\) and
padding 1, reducing the spatial resolution from \(H_l\times W_l\) to
\(\lceil H_l/s_c\rceil \times \lceil W_l/s_c\rceil\), while the pointwise
convolution preserves the channel dimension. Unless otherwise stated, we use
\(s_c=2\). The resulting condensed tokens are used as keys and values in the
Cross-Attention Hyper-Adapter, while the learnable queries remain compact and
rank-oriented. The subsequent query probing and target-specific FC decoding
follow Eqs.~\eqref{eq:T_phi}--\eqref{eq:target_heads}.

\subsection{RankNorm and Sequential Low-Rank Assembly}
\label{app:ranknorm_details}

The factorization \(A^{(l)}_{\tau}(x)B^{(l)}_{\tau}(x)\) is scale-ambiguous
because \(AB=(cA)(B/c)\) for any nonzero scalar \(c\). We therefore normalize
each generated rank component before assembly. For each target \(\tau\) and
rank index \(k\),
\begin{equation}
\big(A^{(l)\prime}_{\tau}(x)\big)_{:,k}
=
\frac{\big(A^{(l)}_{\tau}(x)\big)_{:,k}}
{\|\big(A^{(l)}_{\tau}(x)\big)_{:,k}\|_2+\epsilon},
\qquad
\big(B^{(l)\prime}_{\tau}(x)\big)_{k,:}
=
\frac{\big(B^{(l)}_{\tau}(x)\big)_{k,:}}
{\|\big(B^{(l)}_{\tau}(x)\big)_{k,:}\|_2+\epsilon}.
\end{equation}
This per-instance normalization stabilizes the token-to-rank affinities and avoids cross-sample coupling. The residual update is then implemented as
\begin{equation}
\Delta Y^{(l)}_{\tau}
= \alpha\left(X^{(l)} A^{(l)\prime}_{\tau}(x)\right)B^{(l)\prime}_{\tau}(x),
\end{equation}
which requires two low-rank matrix multiplications and avoids forming the equivalent dense \(d_{\mathrm{in}}\times d_{\mathrm{out}}\) dynamic update.

\section{Detailed Experimental Settings}
\label{app:experimental_details}

\subsection{Datasets and Protocols}
\label{app:datasets_protocols}

We follow standard all-in-one image restoration protocols. AIO-3 trains one model on mixed dehazing, deraining, and denoising data and evaluates it on SOTS, Rain100L, and BSD68 with Gaussian noise levels \(\sigma\in\{15,25,50\}\). AIO-5 further includes deblurring and low-light enhancement and is reported using five benchmark entries: SOTS, Rain100L, BSD68 at \(\sigma=25\), GoPro, and LOLv1. For compositional restoration, we follow the CDD-11 protocol, which includes four single degradations, five double compositions, and two triple compositions. All evaluations are blind: the model receives only the degraded image and does not use test-time degradation labels, prompts, or descriptors.

\subsection{Backbone and CEA Configuration}
\label{app:backbone_config}

We use an asymmetric U-shaped Transformer encoder-decoder backbone following
prior all-in-one restorers. The encoder contains three downsampling stages with
\([4,6,6]\) Transformer blocks, followed by a latent stage with 8 blocks. The
decoder contains \([2,4,4]\) Transformer blocks from low to high resolution,
followed by 4 refinement blocks. We report two model scales: CEA-IR-S uses
initial embedding dimension \(C=32\), while CEA-IR uses \(C=48\); in both cases
the channel dimension doubles after each downsampling stage. The attention heads
are \([1,2,4,8]\) across the four resolution levels, the FFN expansion ratio is
2, and convolution/projection biases are disabled.

Unless otherwise stated, CEA modules are inserted into alternating decoder
blocks, starting from the first block of each decoder stage. This gives \(1\),
\(2\), and \(2\) CEA-equipped blocks in the three decoder stages, respectively.
The default injection targets are the attention query and key projections
(\(Q\) and \(K\)). The low-rank bottleneck is \(r=8\), and each hyper-adapter
uses 4-head cross-attention with 16 learnable queries in total, with 8 queries
for \(A\) and 8 queries for \(B\). The CEA placement, injection targets, rank,
and hyper-adapter configuration are kept the same across the two model scales.

\subsection{Optimization and Training Objective}
\label{app:optimization}

All CEA models and matched controlled variants are trained without degradation labels, prompts, or label-based auxiliary objectives. We optimize
\begin{equation}
\mathcal{L}_{\mathrm{total}}
=\mathcal{L}_{\mathrm{rec}}+\lambda_f\mathcal{L}_{\mathrm{fft}},
\qquad
\mathcal{L}_{\mathrm{rec}}=\|\hat{y}-y\|_1,
\end{equation}
with $\lambda_f=0.10$. The Fourier-domain reconstruction regularizer is
\begin{equation}
\mathcal{L}_{\mathrm{fft}}
=\big\|\,|\mathcal{F}(\hat{y})|-|\mathcal{F}(y)|\,\big\|_1,
\end{equation}
where $\mathcal{F}(\cdot)$ denotes FFT. Following recent restoration practice, this term is used as a shared reconstruction regularizer for all CEA models and matched controlled variants. It is not part of the proposed CEA module and should not be interpreted as label-based auxiliary supervision.

We train all CEA models and matched controlled variants using Adam with $(\beta_1,\beta_2)=(0.9,0.999)$, a total batch size of 64, an initial learning rate of $5\times10^{-4}$, and cosine learning-rate decay. AIO models are trained for 120 epochs, while CDD-11 models are trained for 200 epochs. We use random $128\times128$ crops and horizontal/vertical flips for data augmentation.

\paragraph{Training hardware and wall-clock cost.}
All reported training runs were conducted on a single workstation with four NVIDIA RTX 4090 GPUs. For the lightweight CEA-IR-S model, CDD-11 training took approximately 15 hours for 200 epochs. AIO-3 and AIO-5 training took approximately 30--37.5 hours each, about 2--2.5$\times$ the CDD-11 wall-clock time under the same total batch size of 64 and $128\times128$ crop setting. The wall-clock time can vary with dataset I/O, implementation details, and model scale.

\paragraph{Inference measurement.}
Runtime is measured on a single NVIDIA RTX 4090 GPU after 100 warm-up iterations and averaged over 1000 measured runs.

\section{Additional Quantitative Results}
\label{app:additional_quantitative}

\subsection{Full Results on AIO-3}
\label{app:full_aio3}

Table~\ref{tab:app_aio3_full} provides the full per-task AIO-3 comparison corresponding to the compact summary in the main paper.

\begin{table}[htbp]
\centering
\caption{\textbf{Full AIO-3 benchmark.} Quantitative comparison on three degradations following the standard AiO protocol~\citep{potlapalli2023promptir,zhang2023idr}. We report PSNR (dB, $\uparrow$) / SSIM ($\uparrow$) on SOTS (dehazing), Rain100L (deraining), and BSD68 Gaussian denoising with $\sigma\!\in\!\{15,25,50\}$. Methods are grouped by model capacity (\emph{Light} vs.\ \emph{Heavy}). Best results within each group are highlighted.}

\label{tab:app_aio3_full}
\small
\setlength{\tabcolsep}{3pt} % 控制列间距

% 使用 resizebox 强制对齐页面宽度
\resizebox{\linewidth}{!}{
\begin{tabular}{llccccccccccccc}
\toprule

& \multirow{2}{*}{Method} & \multirow{2}{*}{Params.} & \multicolumn{2}{c}{Dehazing} & \multicolumn{2}{c}{Deraining} & \multicolumn{6}{c}{Denoising} & \multicolumn{2}{c}{Average} \\
\cmidrule(lr){4-5} \cmidrule(lr){6-7} \cmidrule(lr){8-13} \cmidrule(lr){14-15}
& & & \multicolumn{2}{c}{SOTS} & \multicolumn{2}{c}{Rain100L} & \multicolumn{2}{c}{BSD68 $\sigma=15$} & \multicolumn{2}{c}{BSD68 $\sigma=25$} & \multicolumn{2}{c}{BSD68 $\sigma=50$} & PSNR & SSIM \\
\midrule

% 用 \multirow{7} 确保 Light 文字在这 7 行中垂直居中，且不带数字前缀
\multirow{7}{*}{\rotatebox[origin=c]{90}{Light}}
& BRDNet  \citep{tian2020brdnet}      & --   & 23.23 & .895 & 27.42 & .895 & 32.26 & .898 & 29.76 & .836 & 26.34 & .693 & 27.80 & .843 \\
& LPNet   \citep{gao2019dynamicdeblur}      & --   & 20.84 & .828 & 24.88 & .784 & 26.47 & .778 & 24.77 & .748 & 21.26 & .552 & 23.64 & .738 \\
& FDGAN    \citep{dong2020fdgan}     & --   & 24.71 & .929 & 29.89 & .933 & 30.25 & .910 & 28.81 & .868 & 26.43 & .776 & 28.02 & .883 \\
& DL \citep{Fan2021DecoupledLearningPIO}           & 2M   & 26.92 & .931 & 32.62 & .931 & 33.05 & .914 & 30.41 & .861 & 26.90 & .740 & 29.98 & .876 \\
& AirNet \citep{li2022airnet}       & 9M   & 27.94 & .962 & 34.90 & .967 & 33.92 & \textbf{.933} & 31.26 & .888 & 28.00 & .797 & 31.20 & .910 \\
& MoCE-IR-S \citep{zamfir2025moceir} & 11M  & 30.94 & .979 & 38.22 & .983 & 34.08 & \textbf{.933} & 31.42 & .888 & 28.16 & .798 & 32.57 & .916 \\
& \textbf{CEA-IR-S (ours)} & 10M & \textbf{31.90} & \textbf{.981} & \textbf{38.62} & \textbf{.984} & \textbf{34.16} & \textbf{.933} & \textbf{31.51} & \textbf{.889} & \textbf{28.27} & \textbf{.803} & \textbf{32.89} & \textbf{.918} \\

\midrule

% 用 \multirow{8} 确保 Heavy 文字在这 8 行中垂直居中
\multirow{11}{*}{\rotatebox[origin=c]{90}{Heavy}}
& MPRNet \citep{zamir2021mprnet}       & 16M   & 25.28 & .955 & 33.57 & .954 & 33.54 & .927 & 30.89 & .880 & 27.56 & .779 & 30.17 & .899 \\
& PromptIR \citep{potlapalli2023promptir}     & 36M   & 30.58 & .974 & 36.37 & .972 & 33.98 & .933 & 31.31 & .888 & 28.06 & .799 & 32.06 & .913 \\
& Gridformer \citep{Wang2023GridFormer}   & 34M   & 30.37 & .970 & 37.15 & .972 & 33.93 & .931 & 31.37 & .887 & 28.11 & .801 & 32.19 & .912 \\
& PromptIR+Art \citep{Wu2025art}  & 33M   & 30.83 & .979 & 37.94 & .982 & 34.06 & .934 & 31.42 & .891 & 28.14 & .801 & 32.49 & .917 \\
& DA-CLIP  \citep{luo2024daclip}    & 125M  & 29.46 & .963 & 36.28 & .968 & 30.02 & .821 & 24.86 & .585 & 22.29 & .476 & 28.58    & .763   \\
& UniProcessor \citep{duan2024uniprocessor} & 1002M & 31.66 & .979 & 38.17 & .982 & 34.08 & .935 & 31.42 & .891 & 28.17 & .803 & 32.70 & .918 \\
& LoRA-IR \citep{ai2024lorair} & -- & 30.68 & .961 & 37.75 & .979 & 34.06 & .935 & 31.42 & .891 & 28.18 & .803 & 32.42 & .914 \\
& Res-HAIR \citep{cao2024hair} & 29M & 30.98 & .979 & 38.59 & .983 & 34.16 & .935 & 31.51 & .892 & 28.24 & .803 & 32.70 & .919 \\
& VLU-Net \citep{Zeng2025vlu}  & 35M & 30.71 & .980 & \textbf{38.93} & .984 & 34.13 & .935 & 31.48 & .892 & 28.23 & .804 & 32.70 & .919 \\
& MoCE-IR \citep{zamfir2025moceir}   & 25M   & 31.34 & .979 & 38.57 & .984 & 34.11 & .932 & 31.45 & .888 & 28.18 & .800 & 32.73 & .917 \\
& \textbf{CEA-IR (ours)} & 22M & \textbf{32.33} & \textbf{.983} & 38.92 & \textbf{.985} & \textbf{34.22} & \textbf{.937} & \textbf{31.58} & \textbf{.895} & \textbf{28.32} & \textbf{.809} & \textbf{33.07} & \textbf{.922} \\

\bottomrule
\end{tabular}
}
\end{table}

\subsection{Full Results on AIO-5}
\label{app:full_aio5}

Table~\ref{tab:app_aio5_full} provides the full per-task AIO-5 comparison corresponding to the compact summary in the main paper.

\begin{table}[htbp]
\centering
\caption{\textbf{Full AIO-5 benchmark.} Comparison on five degradations under the same AiO training protocol~\citep{potlapalli2023promptir,zhang2023idr}. We evaluate on SOTS (dehazing), Rain100L (deraining), BSD68 ($\sigma\!=\!25$), GoPro (deblurring), and LOLv1 (low-light). PSNR (dB, $\uparrow$) / SSIM ($\uparrow$) are computed on full RGB images. Our method consistently achieves the best average across two model scales.}
\label{tab:app_aio5_full}
\small

\resizebox{\linewidth}{!}{%
\begin{tabular}{>{\centering\arraybackslash}m{1.6em} l c cc cc cc cc cc cc}
\toprule
& Method & Params.
& \multicolumn{2}{c}{Dehazing}
& \multicolumn{2}{c}{Deraining}
& \multicolumn{2}{c}{Denoising}
& \multicolumn{2}{c}{Deblurring}
& \multicolumn{2}{c}{Low-Light}
& \multicolumn{2}{c}{Average} \\
\cmidrule(lr){4-5}\cmidrule(lr){6-7}\cmidrule(lr){8-9}\cmidrule(lr){10-11}\cmidrule(lr){12-13}\cmidrule(lr){14-15}
& & &
\multicolumn{2}{c}{SOTS} &
\multicolumn{2}{c}{Rain100L} &
\multicolumn{2}{c}{BSD68~$\sigma$=25} &
\multicolumn{2}{c}{GoPro} &
\multicolumn{2}{c}{LOLv1} &
PSNR & SSIM \\
\midrule

% ---------------- Light (5 rows) ----------------
\multirow{6}{*}{\rotatebox[origin=c]{90}{Light}}
& SwinIR \citep{liang2021swinir}        & 1M  & 21.50 & .891 & 30.78 & .923 & 30.59 & .868 & 24.52 & .773 & 17.81 & .723 & 25.04 & .835 \\
& DL \citep{Fan2021DecoupledLearningPIO}             & 2M  & 20.54 & .826 & 21.96 & .762 & 23.09 & .745 & 19.86 & .672 & 19.83 & .712 & 21.05 & .743 \\
& TAPE \citep{liu2022tape}           & 1M  & 22.16 & .861 & 29.67 & .904 & 30.18 & .855 & 24.47 & .763 & 18.97 & .621 & 25.09 & .801 \\
& AirNet \citep{li2022airnet}         & 9M  & 21.04 & .884 & 32.98 & .951 & 30.91 & .882 & 24.35 & .781 & 18.18 & .735 & 25.49 & .847 \\
& MoCE-IR-S  \citep{zamfir2025moceir}   & 11M & \textbf{31.33} & .978 & 37.21 & .978 & 31.25 & .884 & 28.90 & .877 & 21.68 & .851 & 30.08 & .913 \\
& \textbf{CEA-IR-S (ours)} & 10M & 31.02 & \textbf{.979} & \textbf{38.50} & \textbf{.984} & \textbf{31.45} & \textbf{.888} & \textbf{29.47} & \textbf{.886} & \textbf{23.58} & \textbf{.858} & \textbf{30.80} & \textbf{.919} \\
\midrule

% ---------------- Heavy (9 rows) ----------------
\multirow{12}{*}{\rotatebox[origin=c]{90}{Heavy}}
& NAFNet \citep{chen2022nafnet}         & 17M & 25.23 & .939 & 35.56 & .967 & 31.02 & .883 & 26.53 & .808 & 20.49 & .809 & 27.76 & .881 \\
& DGUNet   \citep{Mou2022DGUNet}     & 17M & 24.78 & .940 & 36.62 & .971 & 31.10 & .883 & 27.25 & .837 & 21.87 & .823 & 28.32 & .891 \\
& Restormer \citep{zamir2022restormer}     & 26M & 24.09 & .927 & 34.81 & .962 & 31.49 & .884 & 27.22 & .829 & 20.41 & .806 & 27.60 & .881 \\
& MambaIR \citep{Guo2024MambaIR}        & 27M & 25.81 & .944 & 36.55 & .971 & 31.41 & .884 & 28.61 & .875 & 22.49 & .832 & 28.97 & .901 \\
& Transweather \citep{valanarasu2022transweather}   & 38M & 21.32 & .885 & 29.43 & .905 & 29.00 & .841 & 25.12 & .757 & 21.21 & .792 & 25.22 & .836 \\
& IDR  \citep{zhang2023idr}           & 15M & 25.24 & .943 & 35.63 & .965 & \textbf{31.60} & .887 & 27.87 & .846 & 21.34 & .826 & 28.34 & .893 \\
& Gridformer \citep{Wang2023GridFormer}     & 34M & 26.79 & .951 & 36.61 & .971 & 31.45 & .885 & 29.22 & .884 & 22.59 & .831 & 29.33 & .904 \\
& InstructIR-5D \citep{conde2024instructir}    & 17M & 27.10 & .956 & 36.84 & .973 & 31.40 & .873 & 29.40 & .886 & 23.00 & .836 & 29.55 & .908 \\
& Res-HAIR \citep{cao2024hair} & 29M & 30.62 & .978 & 38.11 & .981 & 31.49 & .891 & 28.52 & .874 & 23.12 & .847 & 30.37 & .914 \\
& VLU-Net \citep{Zeng2025vlu}  & 35M & 30.84 & .980 & 38.54 & .982 & 31.43 & .891 & 27.46 & .840 & 22.29 & .833 & 30.11 & .905 \\
& MoCE-IR  \citep{zamfir2025moceir}     & 25M & 30.48 & .974 & 38.04 & .982 & 31.34 & .887 & \textbf{30.05} & \textbf{.899} & 23.00 & .852 & 30.58 & .919 \\
& \textbf{CEA-IR (ours)} & 22M & \textbf{31.96} & \textbf{.981} & \textbf{38.95} & \textbf{.985} & 31.53 & \textbf{.894} & 29.90 & .894 & \textbf{23.32} & \textbf{.857} & \textbf{31.13} & \textbf{.922} \\
\bottomrule
\end{tabular}%
}
\end{table}

\subsection{Full Per-Category Results on CDD-11}
\label{app:full_cdd11}

Table~\ref{tab:app_cdd11_full} provides the full per-category CDD-11 comparison. The main paper reports group-level averages to save space.

\begin{table}[!htbp]
\centering
\caption{\textbf{Full CDD-11 compositional degradations.} Results on the CDD-11 protocol introduced in OneRestore~\citep{guo2024onerestore}. We report PSNR (dB, $\uparrow$) / SSIM ($\uparrow$) on single, double, and triple degradation compositions (L: low-light, H: haze, R: rain, S: snow). Our method achieves the best overall average and yields consistent gains on challenging haze-related compositions.}
\label{tab:app_cdd11_full}
\scriptsize

% 压缩间距
\setlength{\tabcolsep}{2.2pt}
\renewcommand{\arraystretch}{1.08}

\resizebox{\linewidth}{!}{%
\begin{tabular}{l c cccc ccccc cc c}
\toprule
Method & Params.
& \multicolumn{4}{c}{\textit{CDD11-Single}}
& \multicolumn{5}{c}{\textit{CDD11-Double}}
& \multicolumn{2}{c}{\textit{CDD11-Triple}}
& Avg. \\
\cmidrule(lr){3-6}\cmidrule(lr){7-11}\cmidrule(lr){12-13}
& &
Low (L) & Haze (H) & Rain (R) & Snow (S) &
L+H & L+R & L+S & H+R & H+S &
L+H+R & L+H+S &
\\
\midrule

AirNet \citep{li2022airnet}      & 9M  & 24.83/.778 & 24.21/.951 & 26.55/.891 & 26.79/.919 & 23.23/.779 & 22.82/.710 & 23.29/.723 & 22.21/.868 & 23.29/.901 & 21.80/.708 & 22.24/.725 & 23.75/.814 \\
PromptIR \citep{potlapalli2023promptir}    & 36M & 26.32/.805 & 26.10/.969 & 31.56/.946 & 31.53/.960 & 24.49/.789 & 25.05/.771 & 24.51/.761 & 24.54/.924 & 23.70/.925 & 23.74/.752 & 23.33/.747 & 25.90/.850 \\
WGWSNet  \citep{zhu2023wgws}    & 26M & 24.39/.774 & 27.90/.982 & 33.15/.964 & 34.43/.973 & 24.27/.800 & 25.06/.772 & 24.60/.765 & 27.23/.955 & 27.65/.960 & 23.90/.772 & 23.97/.771 & 26.96/.863 \\
WeatherDiff \citep{ozdenizci2023weatherdiffusion} & 83M & 23.58/.763 & 21.99/.904 & 24.85/.885 & 24.80/.888 & 21.83/.756 & 22.69/.730 & 22.12/.707 & 21.25/.868 & 21.99/.868 & 21.23/.716 & 21.04/.698 & 22.49/.799 \\
OneRestore \citep{guo2024onerestore}  & 6M  & 26.48/.826 & 32.52/.990 & 33.40/.964 & 34.31/.973 & 25.79/.822 & 25.58/.799 & 25.19/.789 & 29.99/.957 & 30.21/.964 & 24.78/.788 & 24.90/.791 & 28.47/.878 \\
MoCE-IR-S \citep{zamfir2025moceir} & 11M & 27.26/.824 & 32.66/.990 & 34.31/.970 & 35.91/.980
                 & 26.24/.817 & 26.25/.800 & 26.04/.793 & 29.93/.964 & 30.19/.970
                 & 25.41/.789 & 25.39/.790 & 29.05/.881 \\
\textbf{CEA-IR-S (ours)} & \textbf{10M}
                 & \textbf{27.47/.827} & \textbf{34.75/.993} & \textbf{34.72/.971} & \textbf{36.49/.982}
                 & \textbf{26.67/.824} & \textbf{26.65/.806} & \textbf{26.58/.802} & \textbf{31.15/.968} & \textbf{31.49/.974}
                 & \textbf{25.87/.797} & \textbf{25.93/.799} & \textbf{29.79/.886} \\
\bottomrule
\end{tabular}%
}
% 恢复默认，避免影响后面的表
\setlength{\tabcolsep}{6pt}
\renewcommand{\arraystretch}{1.0}

\end{table}

\section{Additional Statistical and Ablation Studies}
\label{app:additional_ablations}

\subsection{Paired Bootstrap Analysis on AIO-3 and AIO-5}
\label{app:bootstrap_aio}

We perform paired bootstrap analyses on the lightweight AIO-3 and AIO-5 comparisons to assess whether the improvements are consistent across matched test images. Table~\ref{tab:bootstrap_aio} reports the observed paired differences and bootstrap confidence intervals. For each benchmark entry, we compute the per-image PSNR/SSIM difference between CEA-IR-S and MoCE-IR-S, resample matched image pairs with replacement, and report the mean difference together with the 95\% percentile confidence interval over 10,000 bootstrap resamples. This analysis measures test-set paired significance rather than variability across independent training runs. For AIO-5, the overall statistic is computed as the unweighted mean over five benchmark entries: SOTS, Rain100L, BSD68 at \(\sigma=25\), GoPro, and LOLv1. Because Gaussian noise is generated online for denoising and the full benchmark tables report rounded aggregate scores, the bootstrap mean computed from raw matched per-image differences may not exactly equal the arithmetic difference obtained from the displayed rounded table entries.

\begin{table}[!htbp]
\centering
\caption{\textbf{Paired bootstrap analysis on AIO-3 and AIO-5.}
Differences are computed as CEA-IR-S minus MoCE-IR-S on matched test images. Confidence intervals are estimated from 10,000 paired bootstrap resamples.}
\label{tab:bootstrap_aio}
\small
\begin{tabular}{llcc}
\toprule
Benchmark & Metric & Observed difference & 95\% CI \\
\midrule
AIO-3 & PSNR & +0.3046 & [ +0.2565, +0.3582 ] \\
AIO-3 & SSIM & +0.0019 & [ +0.0017, +0.0022 ] \\
AIO-5 & PSNR & +0.6892 & [ +0.3591, +1.0192 ] \\
AIO-5 & SSIM & +0.0054 & [ +0.0033, +0.0073 ] \\
\bottomrule
\end{tabular}
\end{table}

All confidence intervals are strictly positive. Across 10,000 bootstrap resamples, no resampled mean difference is non-positive for either benchmark, giving an empirical one-sided bootstrap probability of \(p_{\mathrm{boot}} < 10^{-4}\). On AIO-5, CEA-IR-S improves most benchmark entries, with the only negative PSNR difference occurring on SOTS dehazing; the overall paired improvement remains positive.

\subsection{Paired Bootstrap Analysis on CDD-11}
\label{app:bootstrap_cdd11}

To further assess the robustness of the CDD-11 improvement, we perform a
paired bootstrap analysis between CEA-IR-S and MoCE-IR-S. The analysis uses
matched test images: for each image, we first compute the PSNR/SSIM difference
between the two methods, and then resample matched image pairs with replacement.
We run 10,000 bootstrap resamples, and Table~\ref{tab:bootstrap_cdd11} reports
the mean difference together with the 95\% percentile confidence interval.

\begin{table}[!htbp]
\centering
\caption{\textbf{Paired bootstrap analysis on CDD-11.}
Differences are computed as CEA-IR-S minus MoCE-IR-S on matched test images.
Confidence intervals are estimated from 10,000 paired bootstrap resamples.}
\label{tab:bootstrap_cdd11}
\small
\begin{tabular}{lccc}
\toprule
Metric & MoCE-IR-S & CEA-IR-S & Difference with 95\% CI \\
\midrule
PSNR & 29.0490 & 29.7924 & +0.7434 \; [ +0.6692, +0.8205 ] \\
SSIM & 0.8806 & 0.8857 & +0.0051 \; [ +0.0047, +0.0057 ] \\
\bottomrule
\end{tabular}
\end{table}

Both confidence intervals are strictly positive, indicating that the CDD-11
gain is consistent under paired resampling rather than being driven by a small
subset of test images. Across 10,000 bootstrap resamples, no resampled mean
difference is non-positive, giving an empirical one-sided bootstrap probability
of \(p_{\mathrm{boot}} < 10^{-4}\).

%------------------------------------------------%

\subsection{Global Conditioning vs. Spatial Query Probing}
\label{app:global_vs_spatial}

To evaluate whether spatial probing is necessary for factor generation, Table~\ref{tab:app_global_vs_spatial} compares CEA with a Global-Dynamic variant that predicts \(A(x)\) and \(B(x)\) from a global average-pooled feature using an MLP. Both variants use the same backbone, Q/K injection, rank \(r=8\), reconstruction losses, and dense softmax-free dot-product assembly; only the factor generator differs.

\begin{table}[!htbp]
\centering
\caption{\textbf{Global conditioning vs. spatial query probing on CDD-11.}
All variants are trained without degradation labels or auxiliary objectives.}
\label{tab:app_global_vs_spatial}
\small
\begin{tabular}{lcccc}
\toprule
Generator & Single & Double & Triple & Avg. \\
\midrule
GAP + MLP & 32.53/.941 & 27.87/.871 & 25.53/.792 & 29.14/.882 \\
Cross-attention queries (CEA) & \textbf{33.36/.943} & \textbf{28.51/.875} & \textbf{25.90/.798} & \textbf{29.79/.886} \\
\bottomrule
\end{tabular}
\end{table}

Spatial query probing improves the overall average from 29.14/0.882 to 29.79/0.886. The gain is observed across all degradation groups, 
supporting the use of cross-attention queries to collect spatial degradation evidence before generating the low-rank factors.
%-----------------------------------------------------%

% \subsection{Architecture vs. Reconstruction Regularization}
% \label{app:loss_disentangle}

% We separate the architectural contribution of CEA from the Fourier-domain reconstruction regularizer. All variants are trained without degradation labels, prompts, or classification/contrastive auxiliary objectives. The Fourier loss is applied as a shared reconstruction regularizer rather than as a method-specific component.

% \begin{table}[t]
% \centering
% \caption{\textbf{Architecture vs. reconstruction regularization on CDD-11.} The Fourier loss is a shared reconstruction regularizer.}
% \label{tab:app_loss_disentangle}
% \small
% \begin{tabular}{lcc}
% \toprule
% Setting & Reconstruction loss & Avg. PSNR/SSIM \\
% \midrule
% Backbone & $L_1$ & xx/.xxx \\
% CEA & $L_1$ & xx/.xxx \\
% Backbone & $L_1+L_{\mathrm{fft}}$ & 29.00/.884 \\
% CEA & $L_1+L_{\mathrm{fft}}$ & 29.79/.886 \\
% \bottomrule
% \end{tabular}
% \end{table}

% Under matched $L_1+L_{\mathrm{fft}}$ training, CEA improves over the backbone by 0.80 dB in average PSNR. After the $L_1$-only runs finish, this table will further test whether the same architectural gain holds without the Fourier-domain regularizer.

\subsection{RankNorm and Rank Sensitivity}
\label{app:ranknorm_rank}

We further analyze two design choices in the low-rank assembly: the use of RankNorm and the rank \(r\). All variants are evaluated on CDD-11 using the lightweight CEA-IR-S setting with the same backbone, training objective, and \(Q{+}K\) injection.

The left part of Table~\ref{tab:app_ranknorm_rank} shows that removing RankNorm reduces the overall average from 29.79/0.886 to 29.59/0.885. This indicates that RankNorm improves the conditioning of dynamically generated factors, although the model remains effective without it. The right part of Table~\ref{tab:app_ranknorm_rank} studies the rank \(r\). Increasing the rank from \(4\) to \(8\) improves the average PSNR from 29.40 to 29.79, while further increasing the rank to \(16\) slightly decreases performance to 29.70. We therefore use \(r=8\) as the default setting, which provides the best accuracy with negligible parameter overhead.

\begin{table}[!htbp]
\centering
\caption{\textbf{RankNorm and rank sensitivity on CDD-11.}
(a) RankNorm improves the conditioning of dynamically generated low-rank factors. 
(b) Performance saturates at \(r=8\), indicating that a compact residual subspace is sufficient.}
\label{tab:app_ranknorm_rank}
\small
\begin{minipage}[t]{0.42\linewidth}
\centering
\textbf{(a) Effect of RankNorm}\\[0.4em]
\begin{tabular}{lcc}
\toprule
Variant & PSNR \(\uparrow\) & SSIM \(\uparrow\) \\
\midrule
w/o RankNorm & 29.59 & .885 \\
w/ RankNorm & \textbf{29.79} & \textbf{.886} \\
\bottomrule
\end{tabular}
\end{minipage}
\hfill
\begin{minipage}[t]{0.53\linewidth}
\centering
\textbf{(b) Effect of rank \(r\)}\\[0.4em]
\begin{tabular}{cccc}
\toprule
\(r\) & Params. & PSNR \(\uparrow\) & SSIM \(\uparrow\) \\
\midrule
4  & 9.708M & 29.40 & .884 \\
8  & 9.712M & \textbf{29.79} & \textbf{.886} \\
16 & 9.718M & 29.70 & .885 \\
\bottomrule
\end{tabular}
\end{minipage}
\end{table}

\subsection{Ablation on Injection Targets}
\label{app:ablation_injection_targets}

\begin{table}[!htbp]
\centering
\caption{\textbf{Ablation on injection targets (CDD-11).} We report PSNR/SSIM averaged over CDD11-Single, CDD11-Double, CDD11-Triple, and the overall average.}
\label{tab:app_ablation_inject}
\small
\begin{tabular}{lcccc}
\toprule
Injection targets & Single Avg. & Double Avg. & Triple Avg. & Overall Avg. \\
\midrule
None (backbone only) & 32.35/.941 & 27.74/.870 & 25.42/.792 & 29.00/.882 \\
$Q$ only & 32.88/.942 & 28.12/.873 & 25.62/.794 & 29.39/.884 \\
$K$ only & 32.81/.942 & 28.00/.872 & 25.39/.792 & 29.28/.883 \\
$V$ only & 32.73/.942 & 27.99/.871 & 25.51/.792 & 29.26/.882 \\
\textbf{$Q{+}K$ (default)} & \textbf{33.36/.943} & \textbf{28.51/.875} & \textbf{25.90/.798} & \textbf{29.79/.886} \\
$Q{+}K{+}V$ & 32.94/.942 & 28.18/.873 & 25.68/.795 & 29.46/.884 \\
$W_1$ only (FFN-in) & 32.73/.942 & 27.93/.871 & 25.50/.792 & 29.23/.883 \\
\bottomrule
\end{tabular}
\end{table}

Table~\ref{tab:app_ablation_inject} studies where to inject the CEA residual. Modulating \(Q\) and \(K\) gives the best overall result, improving over the backbone-only variant by 0.79 dB in PSNR. All single-target variants improve PSNR over the backbone, but their gains are
smaller than \(Q{+}K\) injection. This suggests that CEA is most effective when it reshapes attention matching rather than only changing value content or feed-forward features. Adding \(V\) on top of \(Q{+}K\) reduces the overall average from 29.79/0.886 to 29.46/0.884. We therefore use \(Q{+}K\) injection as the default accuracy--efficiency trade-off.

\FloatBarrier

\section{Additional Out-of-Distribution Evaluation}
\label{app:ood_eval}

\subsection{Zero-Shot Underwater Enhancement on UIEB}
\label{app:uieb_zero_shot}

Table~\ref{tab:uieb_zero_shot} reports out-of-domain generalization on UIEB under the zero-shot protocol used in recent degradation-agnostic restoration work. Models are trained under the all-in-one restoration setting and directly evaluated on the paired UIEB subset without underwater-specific training or finetuning. This experiment is therefore intended as a cross-domain diagnostic rather than a comparison with underwater-supervised enhancement methods.

\begin{table}[!htbp]
\centering
\caption{\textbf{Zero-shot underwater enhancement on UIEB.}
PSNR/SSIM are evaluated on the paired UIEB subset. All methods are evaluated under a zero-shot restoration setting without underwater-specific training or finetuning.}
\label{tab:uieb_zero_shot}
\small
\begin{tabular}{lcc}
\toprule
Method & PSNR & SSIM \\
\midrule
SwinIR & 15.31 & .740 \\
NAFNet & 15.42 & .744 \\
Restormer & 15.46 & .745 \\
AirNet & 15.46 & .745 \\
IDR & 15.58 & .762 \\
PromptIR & 15.48 & .748 \\
MoCE-IR & 15.91 & .765 \\
MIRAGE-S & 17.29 & .773 \\
\midrule
CEA-IR-S & \textbf{17.31} & \textbf{.774} \\
\bottomrule
\end{tabular}
\end{table}

CEA-IR-S obtains 17.31/0.774 on UIEB without underwater-specific training or finetuning. Compared with zero-shot all-in-one baselines, it improves over MoCE-IR by 1.40 dB and is slightly above MIRAGE-S \citep{ren2026efficient}. Since underwater images are outside the training degradation set, these results should be interpreted as an out-of-domain generalization diagnostic rather than evidence of specialization for underwater enhancement.

\FloatBarrier

\section{Additional Qualitative Results}
\label{app:qualitative_results}

%----------------------------------%

\subsection{Qualitative Comparison on AIO-3}
\label{app:aio3_qual}

\begin{figure}[!htbp]
\centering
\includegraphics[width=\textwidth]{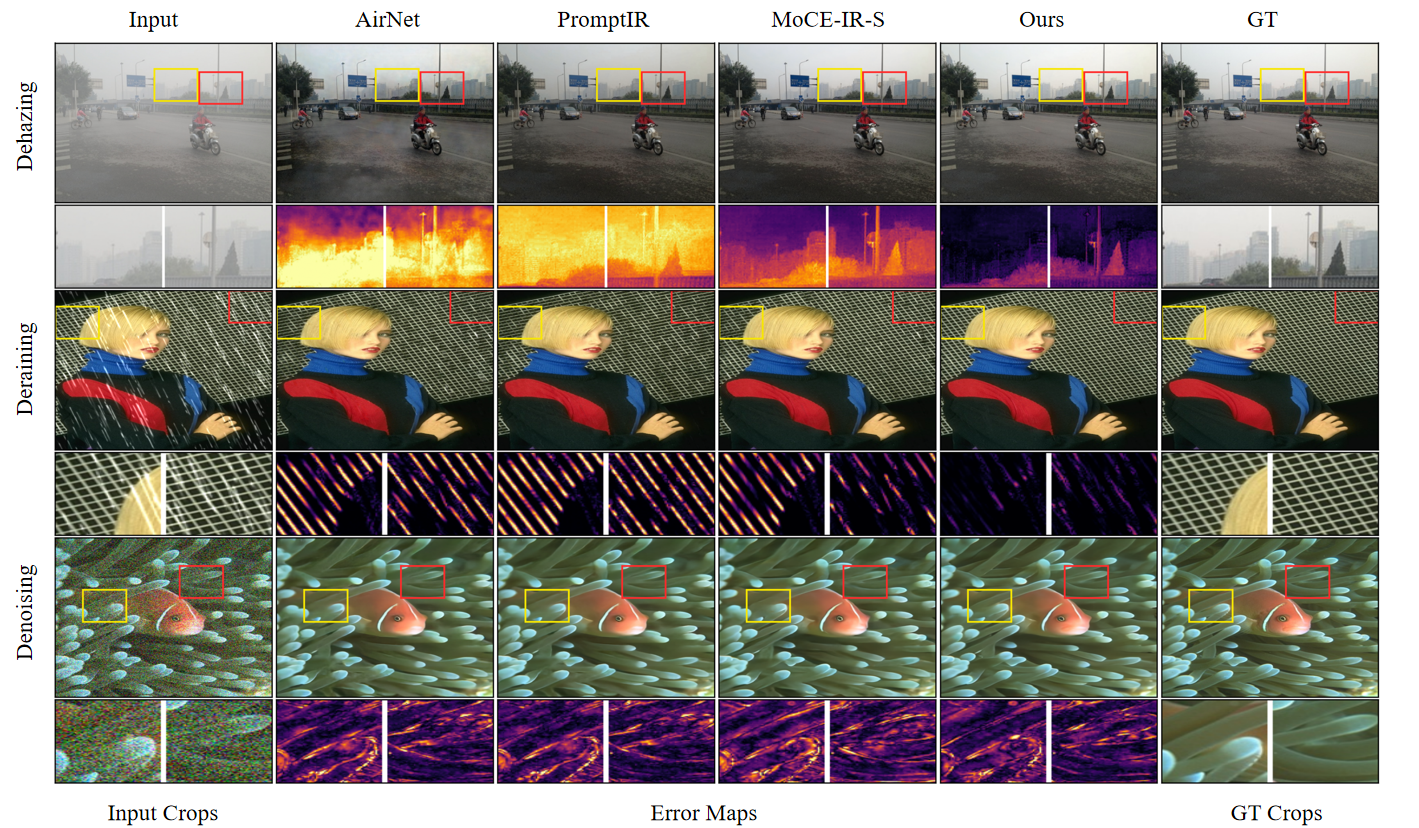}
\caption{\textbf{Qualitative comparison on the AIO-3 benchmark.}
We compare CEA-IR-S against AirNet, PromptIR, and MoCE-IR-S across dehazing, deraining, and denoising. The bottom rows show zoom-in crops and absolute error maps, where darker regions indicate lower reconstruction error. CEA reduces residual degradation artifacts and preserves fine structures across the three restoration tasks.}
\label{fig:app_aio3_qual}
\end{figure}

Figure~\ref{fig:app_aio3_qual} provides representative AIO-3 examples. CEA-IR-S produces fewer residual artifacts in localized degraded regions, such as rain streaks and haze boundaries, while preserving fine structures under denoising.

\FloatBarrier

\subsection{Qualitative Comparison on AIO-5}
\label{app:aio5_qual}

\begin{figure}[!htbp]
\centering
\includegraphics[width=\textwidth]{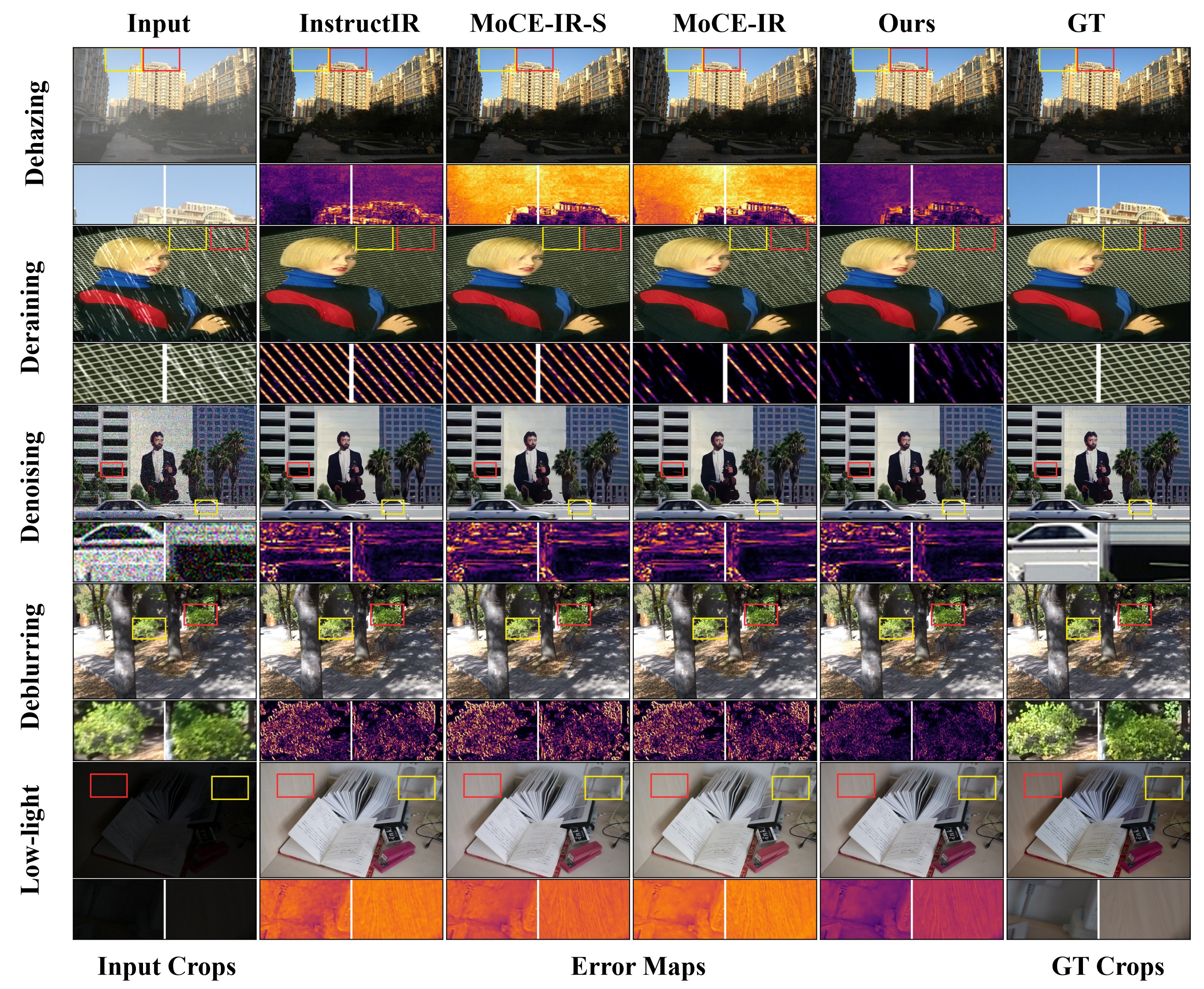}
\caption{\textbf{Additional qualitative comparison on the AIO-5 benchmark.} We compare our method with representative baselines across dehazing, deraining, denoising, deblurring, and low-light enhancement.}
\label{fig:app_aio5_qual}
\end{figure}

Figure~\ref{fig:app_aio5_qual} shows representative AIO-5 examples. Across the five tasks, CEA reduces residual artifacts in local regions and better preserves structural details, while maintaining natural brightness and color in the low-light case.

\FloatBarrier

\subsection{Qualitative Comparison on CDD-11}
\label{app:cdd11_qual}

\begin{figure}[!htbp]
\centering
\includegraphics[width=\textwidth]{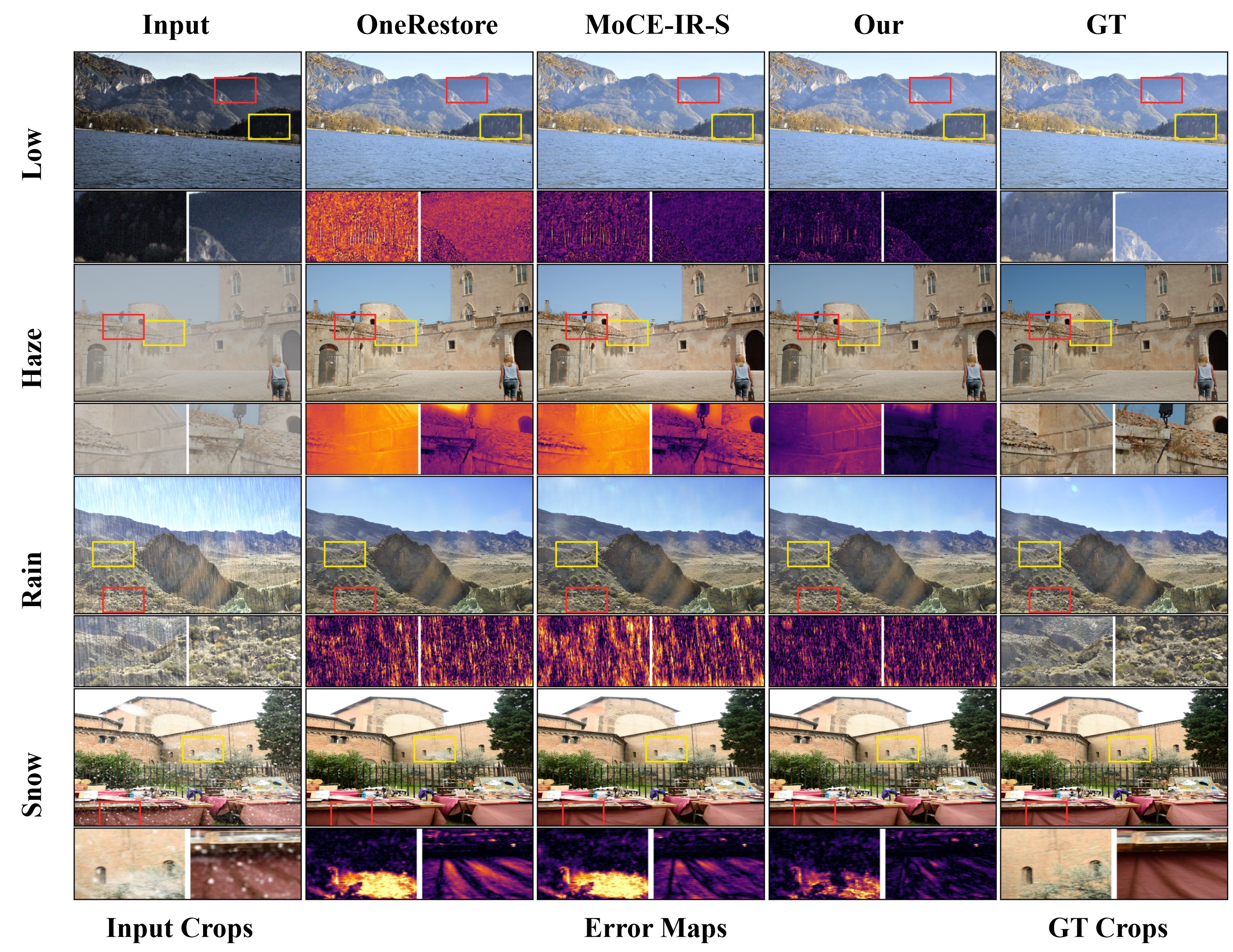}
\caption{\textbf{Qualitative comparison on CDD-11 single degradations.} We compare OneRestore, MoCE-IR-S, and CEA-IR-S on low-light, haze, rain, and snow.}
\label{fig:app_cdd11_single}
\end{figure}

\begin{figure}[!htbp]
\centering
\includegraphics[width=\textwidth]{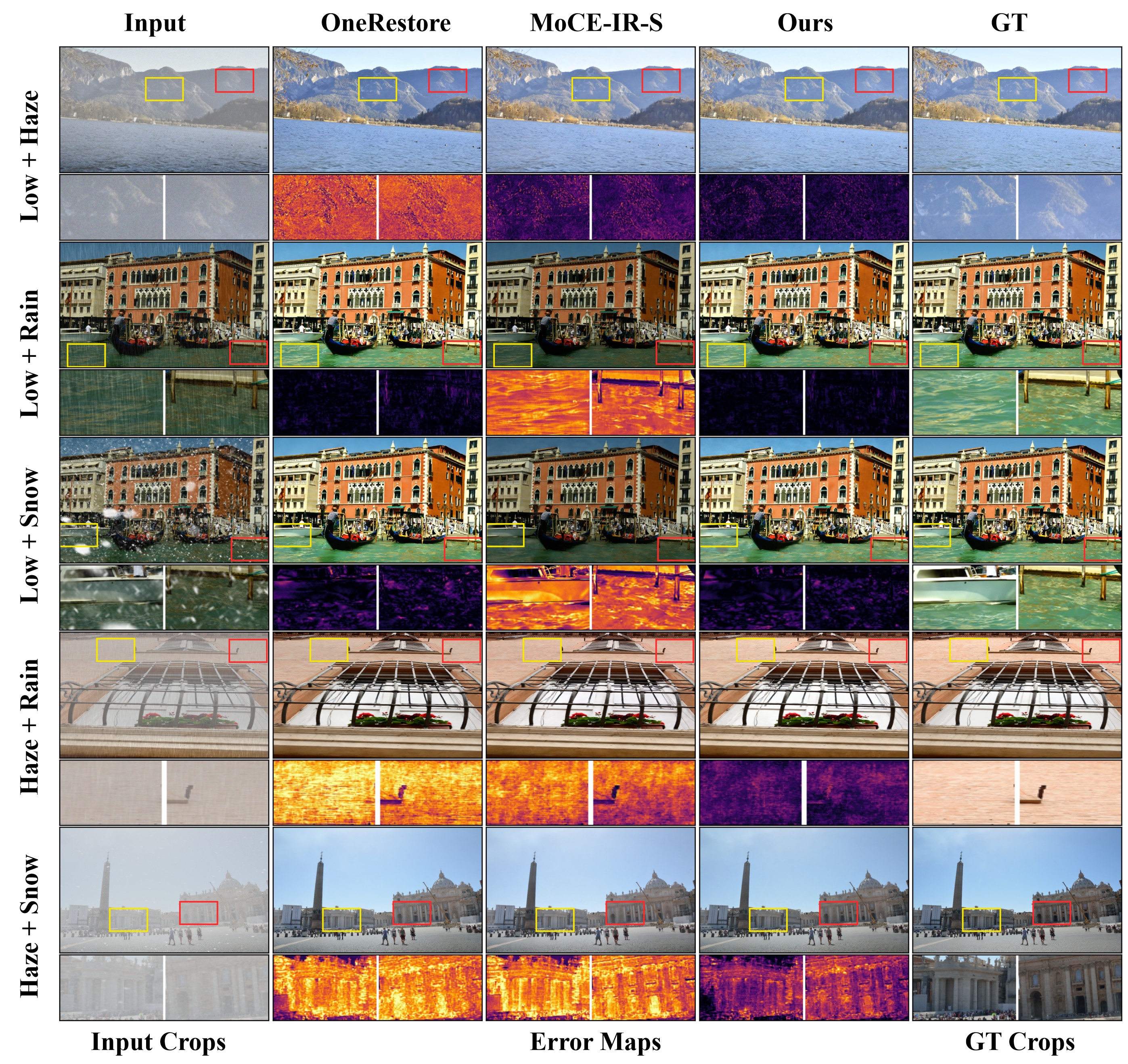}
\caption{\textbf{Qualitative comparison on CDD-11 double degradations.} Representative examples include low-light+haze, low-light+rain, low-light+snow, haze+rain, and haze+snow.}
\label{fig:app_cdd11_double}
\end{figure}

\begin{figure}[!htbp]
\centering
\includegraphics[width=\textwidth]{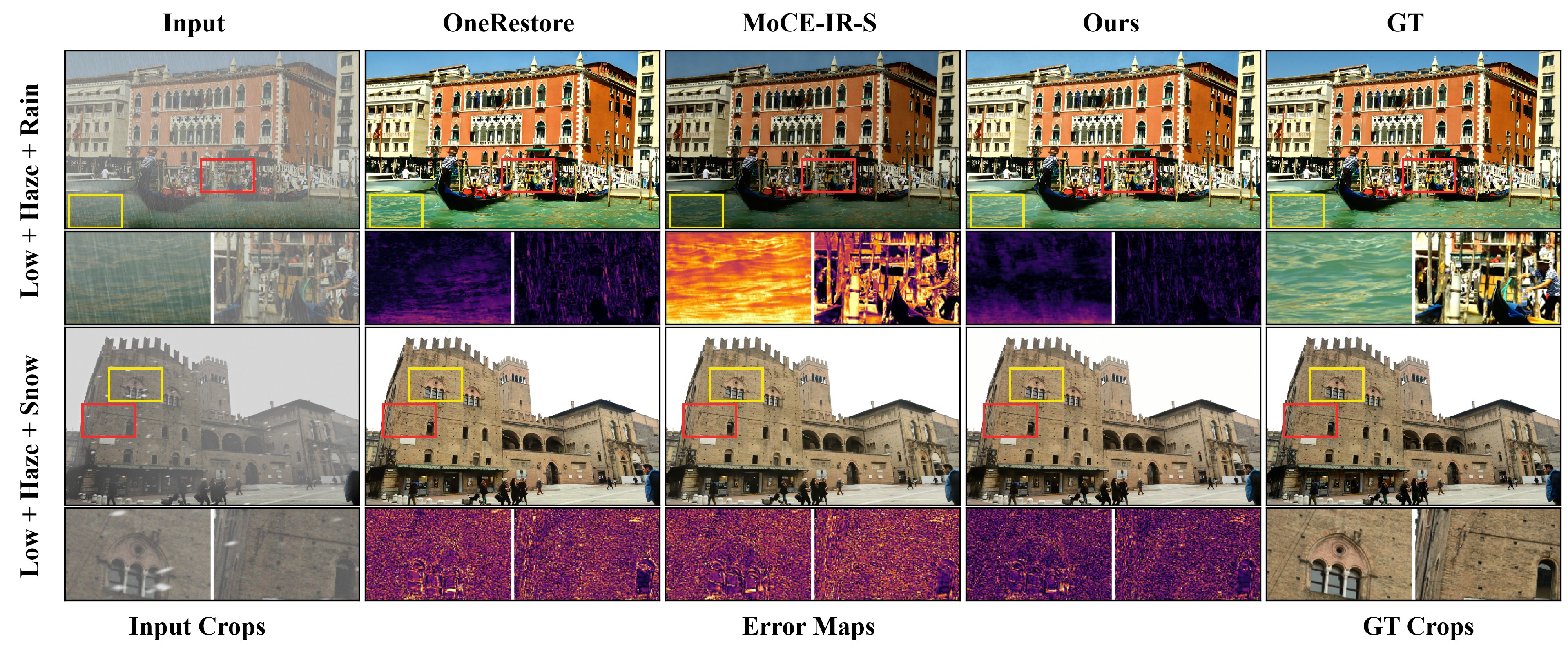}
\caption{\textbf{Qualitative comparison on CDD-11 triple degradations.} We show representative examples with low-light+haze+rain and low-light+haze+snow.}
\label{fig:app_cdd11_triple}
\end{figure}

Figures~\ref{fig:app_cdd11_single}--\ref{fig:app_cdd11_triple} provide additional CDD-11 comparisons. The advantage of CEA is most visible when multiple degradation operators interact, especially in haze-related compositions where both global visibility and local detail preservation are required.

% \begin{minipage}[t]{0.38\textwidth} % 右侧宽度0.38
% \caption{\textbf{Effect of rank $r$.} Performance saturates at $r=8$, indicating an effective information bottleneck.}
% \label{tab:ablate_rank}
% \small

% \begin{tabular*}{\linewidth}{@{\extracolsep{\fill}}cccc@{}}
% \toprule
% $r$ & Params.  & PSNR $\uparrow$ & SSIM $\uparrow$ \\
% \midrule
% 4  &  9.708M & 29.40 & 0.884 \\
% 8  &  9.712M & 29.79 & 0.886 \\
% 16 &  9.718M & 29.70 & 0.885 \\
% \bottomrule
% \end{tabular*}
% \end{minipage}
% \end{table}

%%%%%%%%%%%%%%%%%%%%%%%%%%%%%%%%%%%%%%%%%%%%%%%%%%%%%%%%%%%%

\clearpage

\end{document}